\documentclass[11pt,a4paper]{article}
\usepackage[hyperref]{acl2021}
\usepackage{times}
\usepackage{latexsym}
\usepackage{mathrsfs}
\usepackage{amsmath}
\usepackage{amssymb}
\usepackage{booktabs}
\usepackage{multirow}
\usepackage{graphicx}
\usepackage{multirow}
\usepackage{makecell}
\usepackage{hyperref}
\usepackage{adjustbox}
\usepackage{subcaption}

\usepackage{microtype}

\aclfinalcopy 

\title{X2Parser: Cross-Lingual and Cross-Domain Framework for Task-Oriented Compositional Semantic Parsing}

\author{Zihan Liu, Genta Indra Winata, Peng Xu, Pascale Fung \\
Center for Artificial Intelligence Research (CAiRE)\\
Department of Electronic and Computer Engineering\\
The Hong Kong University of Science and Technology, Clear Water Bay, Hong Kong\\
\texttt{zihan.liu@connect.ust.hk, pascale@ece.ust.hk}}

\date{}

\begin{document}
\maketitle
\begin{abstract}
Task-oriented compositional semantic parsing (TCSP) handles complex nested user queries and serves as an essential component of virtual assistants.
Current TCSP models rely on numerous training data to achieve decent performance but fail to generalize to low-resource target languages or domains.
In this paper, we present \textbf{X2Parser}, a transferable \textbf{Cross}-lingual and \textbf{Cross}-domain \textbf{Parser} for TCSP. Unlike previous models that learn to generate the hierarchical representations for nested intents and slots, we propose to predict flattened intents and slots representations separately and cast both prediction tasks into sequence labeling problems. 
After that, we further propose a fertility-based slot predictor that first learns to dynamically detect the number of labels for each token, and then predicts the slot types.
Experimental results illustrate that our model can significantly outperform existing strong baselines in cross-lingual and cross-domain settings, and our model can also achieve a good generalization ability on target languages of target domains.
Furthermore, our model tackles the problem in an efficient non-autoregressive way that reduces the latency by up to 66\% compared to the generative model.\footnote{The code will be released in \url{https://github.com/zliucr/X2Parser}.}

\end{abstract}

\section{Introduction}
Virtual assistants can perform a wide variety of tasks for users, such as setting reminders, searching for events, and sending messages. 
Task-oriented compositional semantic parsing (TCSP) which comprehends users' intents and detects the key information (slots) in the utterance is one of the core components in virtual assistants.
Existing TCSP models highly rely on large amounts of training data that usually only exist in high-resource domains and languages (e.g., English), and they generally fail to generalize well in a low-resource scenario.
Given that collecting enormous training data is expensive and time-consuming, we aim to develop a transferable model that can quickly adapt to low-resource target languages and domains.

\begin{figure}[t!]
    \centering
    \resizebox{0.47\textwidth}{!}{  
    \includegraphics{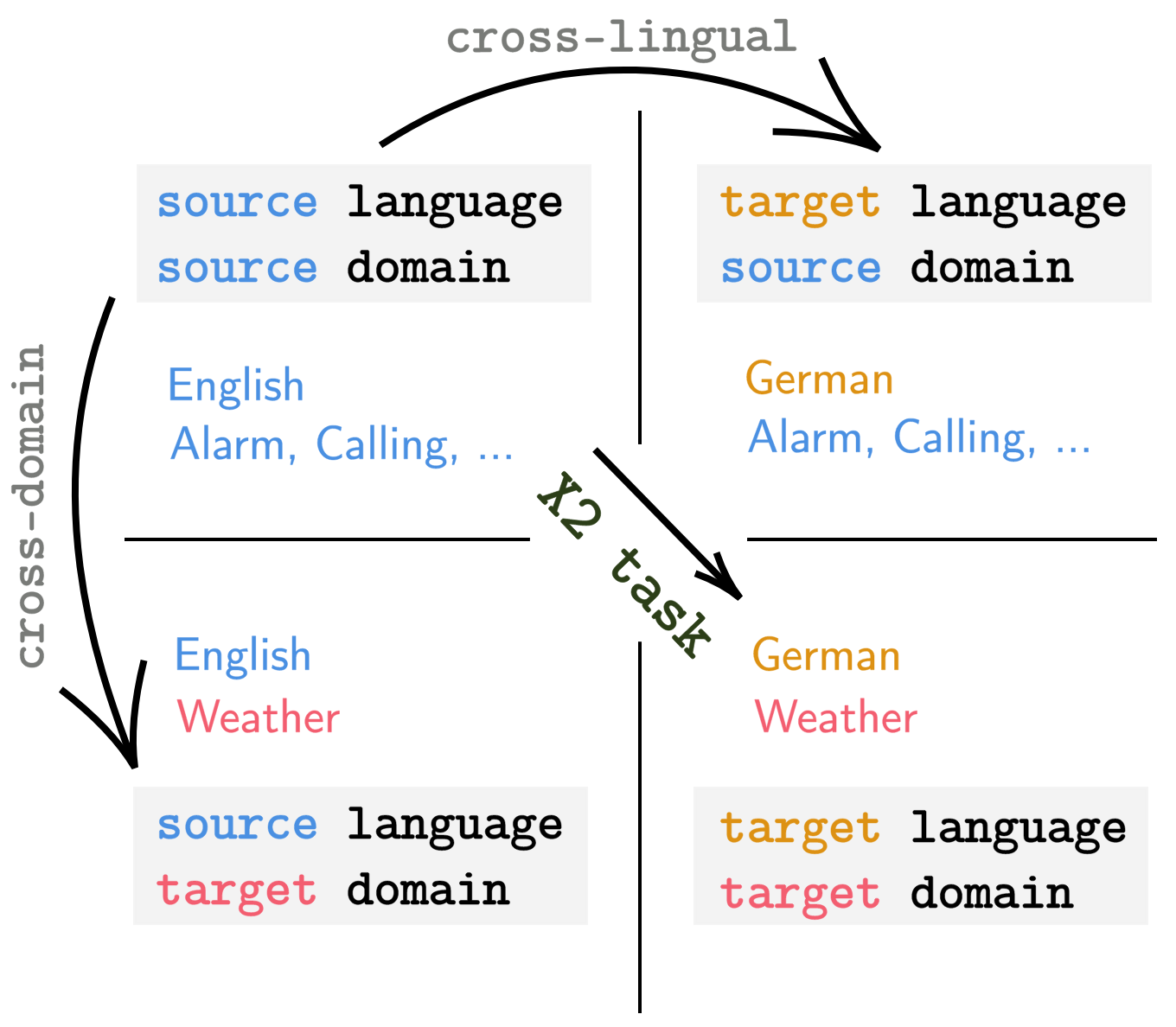}
    }
    \caption{Illustration of the cross-lingual task, cross-domain task, and the combination of both (X2 task).}
    \label{fig:task}
\end{figure}

\begin{figure*}[!ht]
    \centering
    \resizebox{0.999\textwidth}{!}{  
    \includegraphics{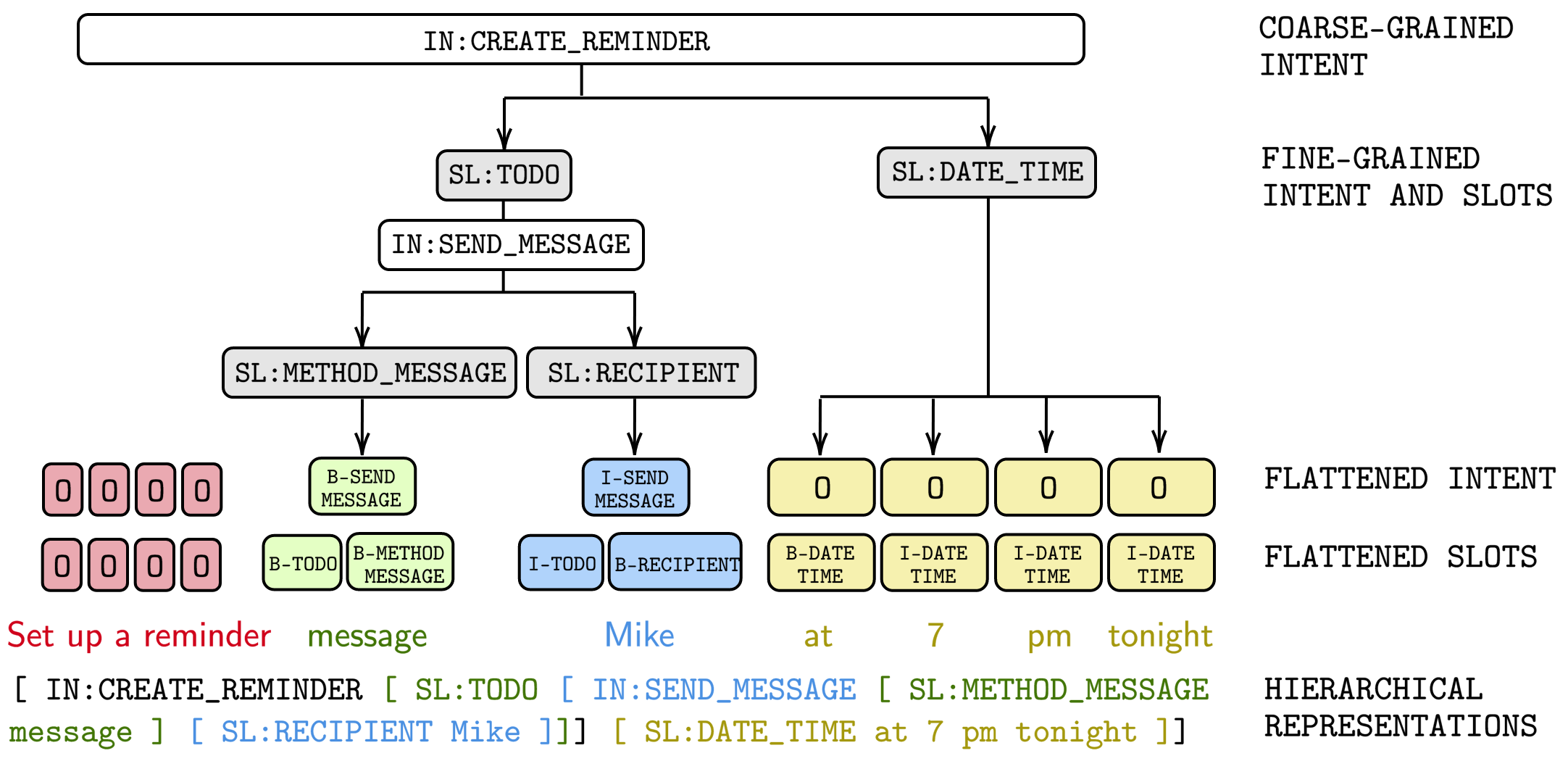}
    }
    \caption{One data example with the illustration of our proposed flattened intents and slots representations, as well as the hierarchical representations used in~\citet{li2020mtop}.}
    \label{fig:example}
\end{figure*}

The traditional semantic parsing can be treated as a simple joint intent detection and slot filling task~\cite{liu2016attention,goo2018slot,zhang2019joint}, while compositional semantic parsing has to cope with complex nested queries, which requires more sophisticated models.
Current state-of-the-art TCSP models~\cite{rongali2020don,li2020mtop} are generation-based models that learn to directly generate the hierarchical representations which contain nested intent and slot labels.\footnote{An example of hierarchical representations is illustrated at the bottom of Figure~\ref{fig:example}.}
We argue that the hierarchical representations are relatively complex, and the models need to learn when to generate the starting intent or slot label, when to copy tokens from the input, and when to generate the end of the label. Hence, large quantities of training data are necessary for the models to learn these complicated skills~\cite{rongali2020don}, while they cannot generalize well when large datasets are absent~\cite{li2020mtop}. Moreover, the inference speed of generation-based models will be greatly limited by the output length.

In this paper, we propose a transferable cross-lingual and cross-domain parser (X2Parser) for TCSP. Instead of generating hierarchical representations, we convert the nested annotations into flattened intent and slot representations (as shown in Figure~\ref{fig:example}) so that the model can learn to predict the intents and slots separately.
We cast the nested slot prediction problem into a special sequence labeling task where each token can have multiple slot labels. To tackle this task, our model first learns to predict the number of slot labels, which helps it capture the hierarchical slot information in user queries. Then, it copies the corresponding hidden state for each token and uses those hidden states to predict the slot labels.
For the nested intent prediction, we cast the problem into a normal sequence labeling problem where each token only has one intent label since the nested cases for intents are simpler than those for slots.
Compared to generation-based models~\cite{li2020mtop}, X2Parser simplifies the problem by flattening the hierarchical representations and tackles the task in a non-autoregressive way, which strengthen its adaptation ability in low-resource scenarios and greatly reduce the latency.

As shown in Figure~\ref{fig:task}, we conduct experiments on three low-resource settings: cross-lingual, cross-domain, and a combination of both. Results show that our model can remarkably surpass existing strong baselines in all the low-resource scenarios by more than 10\% exact match accuracy, and can reduce the latency by up to 66\% compared to generation-based models.
We summarize the main contributions of this paper as follows:
\begin{itemize}
    \item We provide a new perspective to tackle the TCSP task, which is to flatten the hierarchical representations and cast the problem into several sequence labeling tasks.
    \item X2Parser can significantly outperform existing strong baselines in different low-resource settings and notably reduce the latency compared to the generation-based model.
    \item We conduct extensive experiments in different few-shot settings and explore the combination of cross-lingual and cross-domain scenarios. 
    
\end{itemize}

\begin{figure*}[!ht]
    \centering
    \resizebox{0.99\textwidth}{!}{  
    \includegraphics{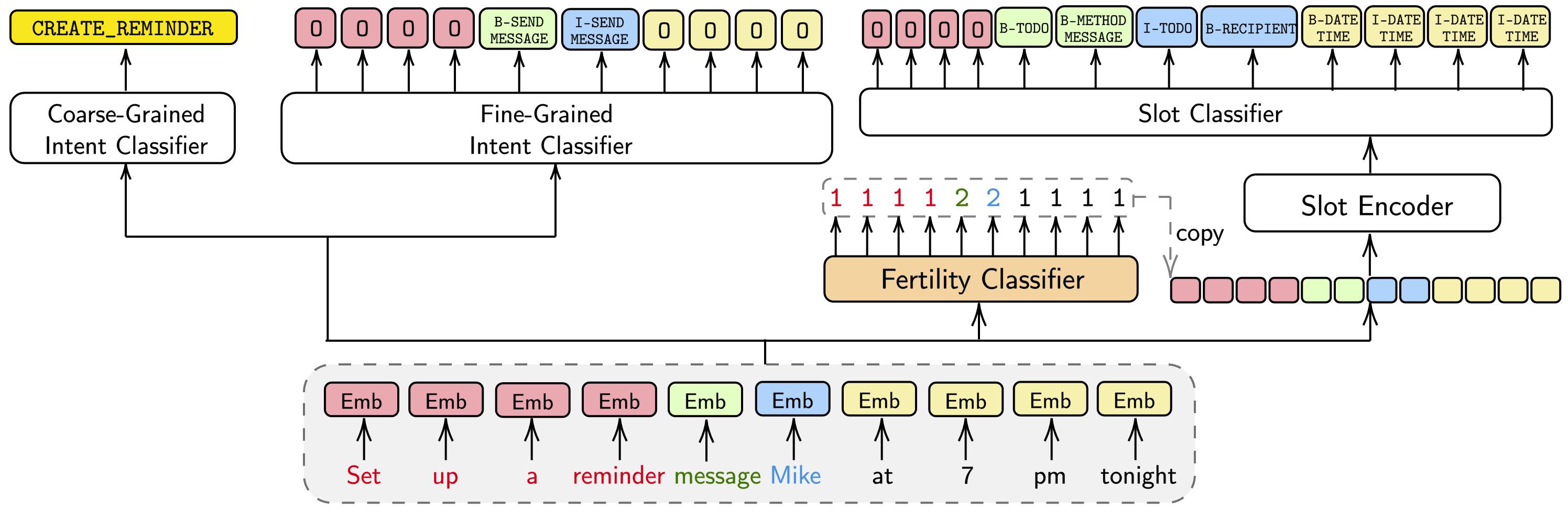}
    }
    \caption{The architecture of X2Parser. We consider the TCSP task as a combination of the coarse-grained intent classification, fine-grained intent prediction, and slot filling tasks.}
    \label{fig:architecture}
\end{figure*}

\section{Related Work}

\subsection{Task-Oriented Semantic Parsing}
The majority of works on task-oriented semantic parsing focused on non-compositional user queries~\cite{mesnil2013investigation,liu2016attention,goo2018slot,zhang2019joint}, which turns the parsing task into a combination of intent detection and slot filling. Recently, \citet{gupta2018semantic} introduced a new dataset, called TOP, annotated with complex nested intents and slots and proposed to use the hierarchical representations to model the task. After that, \citet{rongali2020don} showed that leveraging a sequence-to-sequence model based on a copy mechanism~\cite{see2017get} to directly generate the hierarchical representations was effective at parsing the nested queries.
Taking this further, \citet{chen2020low} and \citet{li2020mtop} extended the TOP dataset into multiple domains and multiple languages, and \citet{li2020mtop} conducted zero-shot cross-lingual experiments using the combination of the multilingual pre-trained models~\cite{conneau2020unsupervised,NEURIPS2020_1763ea5a} and the copy mechanism method proposed in~\citet{rongali2020don}. Lately, \citet{babu2021non} and \citet{shrivastava2021span}, which are concurrent works of X2Parser, proposed to tackled the TCSP task in a non-autoregressive way. Different from them, we propose to flatten the hierarchical representations and cast the problem into several sequence labeling tasks.

\subsection{Language and Domain Adaptation}
Recently, cross-lingual and cross-domain models that aim to tackle low-resource issues have been applied to natural language understanding~\cite{conneau2018xnli,huang2019unicoder,conneau2020unsupervised,gururangan2020don}, sentiment analysis~\cite{zhou2016attention,ziser2017neural}, task-oriented semantic parsing~\cite{chen2018xl,schuster2019cross,liu2019zero,wu2019transferable,liu2020attention,chen2020low,liu2020coach}, named entity recognition~\cite{ni2017weakly,xie2018neural,jia2019cross,liu2020crossner}, speech recognition~\cite{mimura2017cross,winata2020learning}, abstractive summarization~\cite{zhu2019ncls,ouyang2019robust,yu2021adaptsum}, etc. Despite numerous studies related to the cross-lingual and cross-domain areas, only a few of them have explored how to effectively adapt models to the target languages in target domains, and the investigated tasks are limited to sentiment analysis~\cite{fernandez2016distributional,li2020unsupervised}, abusive language detection~\cite{pamungkas2019cross}, and machine reading comprehension~\cite{charlet2020cross}. To the best of our knowledge, we are the first to study the combination of cross-lingual and cross-domain adaptations in the TCSP task.

\section{Task Decomposition}
In this section, we first introduce the intuition of decomposing the compositional semantic parsing into intent predictions and slot filling. Then, we describe how we construct intent and slot labels.

\subsection{Intuition of Task Decomposition}
We argue that hierarchical representations containing nested annotations for intents and slots are relatively complex. We need large enough training data to train a good model based on such representations, and the model's performance will be greatly limited in low-resource scenarios. 
Therefore, instead of incorporating intents and slots into one representation, we propose to predict them separately so that we can simplify the parsing problem and enable the model to easily learn the skills for each decomposed task, and finally, our model can achieve a better adaptation ability in low-resource scenarios.
As illustrated in Figure~\ref{fig:example}, we obtain the coarse-grained intent, flattened fine-grained intents and flattened slot labels from the hierarchical representations, and train the model based on these three categories in a multi-task fashion. Note that we can always reconstruct the hierarchical representations based on the labels in these three categories, which means that the decomposed labels and the hierarchical labels are equivalent. 

\subsection{Label Constructions}
\paragraph{Slot Labels} We extract nested slot labels from the hierarchical representations and assign the labels to corresponding tokens based on the BIO (begin-inside-outside) structure. As we can see from Figure~\ref{fig:example}, there could exist multiple slot labels for one token, and we consider the order of the labels so as to reconstruct the hierarchical representations. Specifically, we put the more fine-grained slot label at the later position. For example, ``message'' (in Figure~\ref{fig:example}) has \texttt{B-TODO} and \texttt{B-METHOD-MESSAGE} labels, and \texttt{B-METHOD-MESSAGE} comes after \texttt{B-TODO} since it is a more fine-grained slot label.

\paragraph{Intent Labels} Each data sample has one intent label for the whole user utterance, and we extract it as an individual coarse-grained intent label. For the intents expressed by partial tokens (i.e., fine-grained intents), we use the BIO structure to label the corresponding tokens. We notice that we only need to assign one intent label to each token since the nested cases for intents are relatively simple.\footnote{We place more details about how we construct labels for fine-grained nested intents in the Appendix~\ref{Appendix_A}.} Therefore, the fine-grained intent classification becomes a  sequence labeling task.

\section{X2Parser}
The model architecture of our X2Parser is illustrated in Figure~\ref{fig:architecture}. To enable the cross-lingual ability of our model, we leverage the multilingual pre-trained model XLM-R~\cite{conneau2020unsupervised} as the sequence encoder.
Let us define $ X = \{x_1, x_2, ..., x_n\} $ as the user utterance and $ H = \{h_1, h_2, ..., h_n\}$ as the hidden states (denoted as Emb in Figure~\ref{fig:architecture}) from XLM-R.

\subsection{Slot Predictor}
The slot predictor consists of a fertility classifier, a slot encoder, and a slot classifier.
Inspired by~\citet{gu2018nonautoregressive}, the fertility classifier learns to predict the number of slot labels for each token, and then it copies the corresponding number of hidden states. Finally, the slot classifier is trained to conduct the sequence labeling based on the slot labels we constructed. 
The fertility classifier not only helps the model identify the number of labels for each token but also guides the model to implicitly learn the nested slot information in user queries. It relieves the burden of the slot classifier, which needs to predict multiple slot entities for certain tokens.

\paragraph{Fertility Classifier (FC)}
We add a linear layer (FC) on top of the hidden states from XLM-R to predict the number of labels (fertility), which we formulate as follows:
\begin{equation}
    F = \{f_1, f_2, ..., f_n\} = \text{FC}(\{h_1, h_2, ..., h_n\}),
\end{equation}
where FC is an n-way classifier (n is the maximum label number) and $f_i (i \in [1,n])$ is a positive integer representing the number of labels for $x_i$. 

\paragraph{Slot Filling}
After obtaining the fertility predictions, we copy the corresponding number of hidden states from XLM-R:
\begin{equation}
    H' = \text{CopyHiddens}(H, F).
\end{equation}
Then, we add a transformer encoder~\cite{vaswani2017attention} (slot encoder (SE)) on top of $H'$ to incorporate the sequential information into the hidden states, followed by adding a linear layer (slot classifier (SC)) to predict the slots, which we formulate as follows:
\begin{equation}
    P_{\text{slot}} = \text{SC}(\text{SE}(H')),
\end{equation}
where $P_{\text{slot}}$ is a sequence of slots that has the same length as the sum of the fertility numbers.

\subsection{Intent Predictor}

\paragraph{Coarse-Grained Intent} 
The coarse-grained intent is predicted based on the hidden state of the ``\texttt{[CLS]}'' token from XLM-R since it can be the representation for the whole sequence, and then we add a linear layer (coarse-grained intent classifier (CGIC)) on top of the hidden state to predict the coarse-grained intent:
\begin{equation}
    p_{\text{cg}} = \text{CGIC}(h_{\text{cls}}),
\end{equation}
where $ p_{\text{cg}} $ is a single intent prediction.

\paragraph{Fine-Grained Intent}
We add a linear layer (fine-grained intent classifier (FGIC)) on top of the hidden states H to produce the fine-grained intents:
\begin{equation}
    P_{\text{fg}} = \text{FGIC}(\{h_1, h_2, ..., h_n\}),
\end{equation}
where $P_{\text{fg}}$ is a sequence of intent labels that has the same length as the input sequence.

\begin{table*}[!th]
\renewcommand{\arraystretch}{1.3}
\centering
\begin{adjustbox}{width={0.95\textwidth},totalheight={\textheight},keepaspectratio}
\begin{tabular}{p{7cm}|cccccc|c}
\Xhline{2\arrayrulewidth}
\textbf{Model}   & \textbf{en}     & \textbf{es} & \textbf{fr} & \textbf{de} & \textbf{hi} & \textbf{th} & \textbf{Avg.} \\ \hline
Seq2Seq w/ CRISS~\cite{li2020mtop}  & \textbf{84.20}        & 48.60       & 46.60       & 36.10       & 31.20       & 0.00        & 32.50         \\
Seq2Seq w/ XLM-R~\cite{li2020mtop}    & 83.90       & 50.30       & 43.90       & 42.30       & 30.90       & 26.70       & 38.82         \\
Neural Layered Model (NLM) & 82.40 & 59.99       & 58.16       & 54.91       & 29.31       & 28.78       & 46.23         \\ \hline
X2Parser        & 83.39           & \textbf{60.30}       & \textbf{58.34}       & \textbf{56.16}       & \textbf{37.06}       & \textbf{29.35}       & \textbf{48.24}         \\ \Xhline{2\arrayrulewidth}
\end{tabular}
\end{adjustbox}
\caption{Exact match accuracies for the zero-shot \textbf{cross-lingual setting}. ``Avg.'' denotes the averaged performance over all target languages (English excluded). The results of X2Parser and NLM are averaged over five runs.}
\label{tab:cross_lingual_results}
\end{table*}

\begin{table*}[!ht]
\renewcommand{\arraystretch}{1.3}
\centering
\begin{adjustbox}{width={0.999\textwidth},totalheight={\textheight},keepaspectratio}
\begin{tabular}{ccccccccccccc}
\Xhline{2\arrayrulewidth}
\multicolumn{1}{c|}{\textbf{Model}}    & \textbf{Alarm} & \textbf{Call.} & \textbf{Event} & \textbf{Msg.}  & \textbf{Music} & \textbf{News}  & \textbf{People} & \textbf{Recipe} & \textbf{Remind} & \textbf{Timer} & \multicolumn{1}{c|}{\textbf{Weather}} & \textbf{Avg.}  \\ \hline
\multicolumn{1}{c|}{Seq2Seq}  & 67.94   & 64.25      & 61.93      & 50.11      & 32.20   & 43.20      & 52.54       & 34.21       & 46.32       & 44.83      & \multicolumn{1}{c|}{73.58}        &  51.92     \\
\multicolumn{1}{c|}{NLM}      & 76.32 & 70.02 & 73.60 & 70.58 & \textbf{56.52} & 58.01 & 67.33  & 50.01  & 57.28  & 64.37 & \multicolumn{1}{c|}{80.15}   & 65.83 \\ \hline
\multicolumn{1}{c|}{X2Parser} & \textbf{76.72} & \textbf{73.16} & \textbf{77.33} & \textbf{71.45} & 55.19 & \textbf{64.43} & \textbf{69.77}  & \textbf{51.78}  & \textbf{58.86}  & \textbf{65.98} & \multicolumn{1}{c|}{\textbf{81.17}}   & \textbf{67.80} \\ \Xhline{2\arrayrulewidth}
\end{tabular}
\end{adjustbox}
\caption{Exact match accuracies (averaged over three runs) for the \textbf{cross-domain setting} in English. The scores represent the performance for the corresponding target domains. We use 10\% of training samples in the target domain. \textbf{``Seq2Seq'' denotes the ``Seq2Seq w/ XLM-R'' baseline} (same for the following tables and figures).}
\label{tab:cross_domain_results}
\end{table*}

\section{Experiments}

\subsection{Experimental Setup}
\paragraph{Dataset}
We conduct the experiments on the MTOP dataset proposed by~\citet{li2020mtop}, which contains six languages: English (en), German (de), French (fr), Spanish (es), and Thai (th), and 11 domains: alarm, calling, event, messaging, music, news, people, recipes, reminder, timer, and weather. The data statistics are reported in the Appendix~\ref{Appendix_B}.

\paragraph{Cross-Lingual Setting}
In the cross-lingual setting, we use English as the source language and the other languages as target languages. In addition, we consider a zero-shot scenario where we only use English data for training.

\paragraph{Cross-Domain Setting}
In the cross-domain setting, we only consider training and evaluation in English.
We choose ten domains as source domains and the other domain as the target domain. Different from the cross-lingual setting, we consider a few-shot scenario where we first train the model using the data from the ten source domains, and then we fine-tune the model using a few data samples (e.g., 10\% of the data) from the target domain.
We consider the few-shot scenario because zero-shot adapting the model to the target domain is extremely difficult due to the unseen intent and slot types, while zero-shot to target languages is easier using multilingual pre-trained models.

\paragraph{Cross-Lingual Cross-Domain Setting} 
This setting combines the cross-lingual and cross-domain settings. Specifically, we first train the model on the English data from the ten source domains, and then fine-tune it on a few English data samples from the other (target) domain. Finally, we conduct the zero-shot evaluation on all the target languages of the target domain.

\begin{table*}[!ht]
\renewcommand{\arraystretch}{1.3}
\centering
\begin{adjustbox}{width={0.999\textwidth},totalheight={\textheight},keepaspectratio}
\begin{tabular}{ccccccccccccc}
\Xhline{2\arrayrulewidth}
\multicolumn{1}{c|}{\textbf{Model}}    & \textbf{Alarm} & \textbf{Call.} & \textbf{Event} & \textbf{Msg.}  & \textbf{Music} & \textbf{News}  & \textbf{People} & \textbf{Recipe} & \textbf{Remind} & \textbf{Timer} & \multicolumn{1}{c|}{\textbf{Weather}} & \textbf{Avg.}  \\ \hline
\multicolumn{1}{c|}{Seq2Seq}  & 34.29      &  47.00     & 41.81      & 25.86      & 19.21      & 25.39      &  22.13      & 16.12       & 9.80       &  20.01     & \multicolumn{1}{c|}{36.90}        &  22.25     \\
\multicolumn{1}{c|}{NLM}      & 48.53 & 43.30 & 44.62 & 43.32 & 36.25 & 28.60 & 43.29  & 28.54  & 20.50  & 34.16 & \multicolumn{1}{c|}{59.57}   & 39.15 \\ \hline
\multicolumn{1}{c|}{X2Parser} & \textbf{48.72} & \textbf{51.30} & \textbf{53.22} & \textbf{43.99} & \textbf{37.25} & \textbf{34.85} & \textbf{45.97}  & \textbf{32.99}  & \textbf{27.87}  & \textbf{36.61} & \multicolumn{1}{c|}{\textbf{60.05}}   & \textbf{42.98} \\ \Xhline{2\arrayrulewidth}
\end{tabular}
\end{adjustbox}
\caption{Exact match accuracies (averaged over three runs) for the \textbf{cross-lingual cross-domain setting}. The result for each domain is the averaged performance over all target languages. We use 10\% of training samples in the English target domain, and do not use any data in the target languages.}
\label{tab:cross_lingual_cross_domain_results}
\end{table*}

\begin{figure*}[!ht]
    \centering
    \resizebox{\textwidth}{!}{  
    \includegraphics{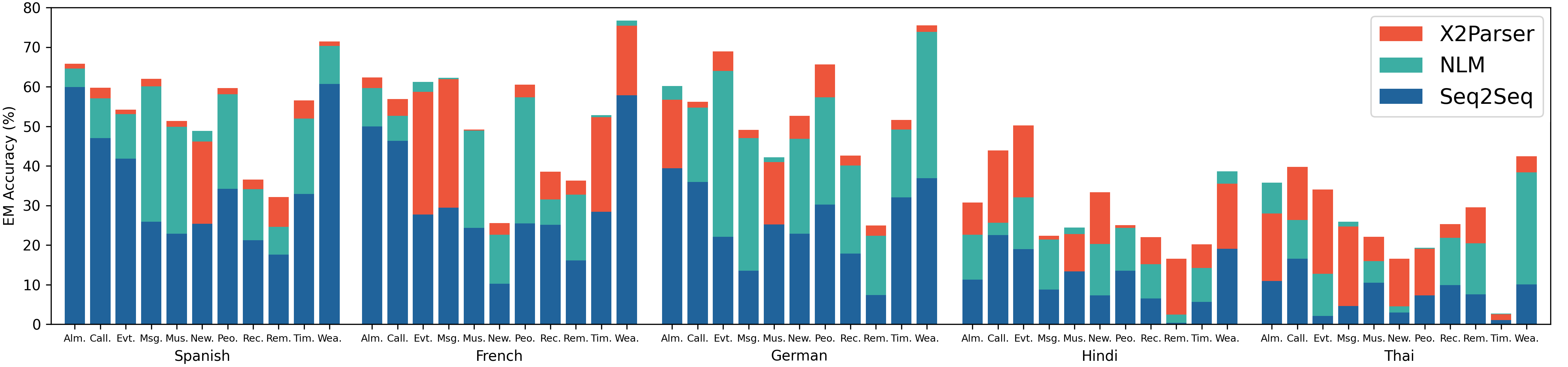}
    }
    \caption{Full \textbf{cross-lingual cross-domain} results (across all target languages of target domains) for Table~\ref{tab:cross_lingual_cross_domain_results}.}
    \label{fig:results-combined}
\end{figure*}

\subsection{Baselines}

\paragraph{Seq2Seq w/ XLM-R} \citet{rongali2020don} proposed a sequence-to-sequence (Seq2Seq) model using a pointer-generator network~\cite{see2017get} to handle nested queries, and achieved new state-of-the-art results in English. \citet{li2020mtop} adopted this architecture for zero-shot cross-lingual adaptation. They replaced the encoder with the XLM-R~\cite{conneau2020unsupervised} and used a customized decoder to learn to generate intent and label types and copy tokens from the inputs.\footnote{In order to compare the performance in the cross-domain and cross-lingual cross-domain settings, we follow~\citet{li2020mtop} to reimplement this baseline since the source code is not publicly available.}

\paragraph{Seq2Seq w/ CRISS}
It is the same architecture as \textit{Seq2Seq w/ XLM-R}, except that \citet{li2020mtop} replaced XLM-R with the multilingual pre-trained model, CRISS~\cite{NEURIPS2020_1763ea5a}, as the encoder for the zero-shot cross-lingual adaptation.

\paragraph{Neural Layered Model (NLM)}
This baseline conducts the multi-task training based on the same task decomposition as X2Parser, but it replaces the slot predictor module in X2Parser with a neural layered model~\cite{ju2018neural},\footnote{This model was originally proposed to tackle the nested named entity recognition task} while keeping the other modules the same.
Unlike our fertility-based slot predictor, NLM uses several stacked layers to predict entities of different levels. We use this baseline to verify the effectiveness of our fertility-based slot predictor.

\subsection{Training Details}
We use XLM-R Large~\cite{conneau2020unsupervised} as the sequence encoder. For a word (in an utterance) with multiple subword tokens, we take the representations from the first subword token to predict the labels for this word. The transformer encoder (slot encoder) has one layer with a head number of 4, a hidden dimension of 400, and a filter size of 64. We set the fertility classifier as a 3-way classifier since the maximum label number for each token in the dataset is 3. We train X2Parser using the Adam optimizer~\cite{kingma2015adam} with a learning rate of 2e-5 and a batch size of 32.
We follow~\citet{li2020mtop} and use the exact match accuracy to evaluate the models. For our model, the prediction is considered correct only when the predictions for the coarse-grained intent, fine-grained intents, and the slots are all correct. To ensure a fair comparison, we use the same three random seeds to run each model and calculate the averaged score for each target language and domain.

\begin{figure*}[!t]
\begin{minipage}{.33\textwidth}
  \centering
  \includegraphics[width=1\linewidth]{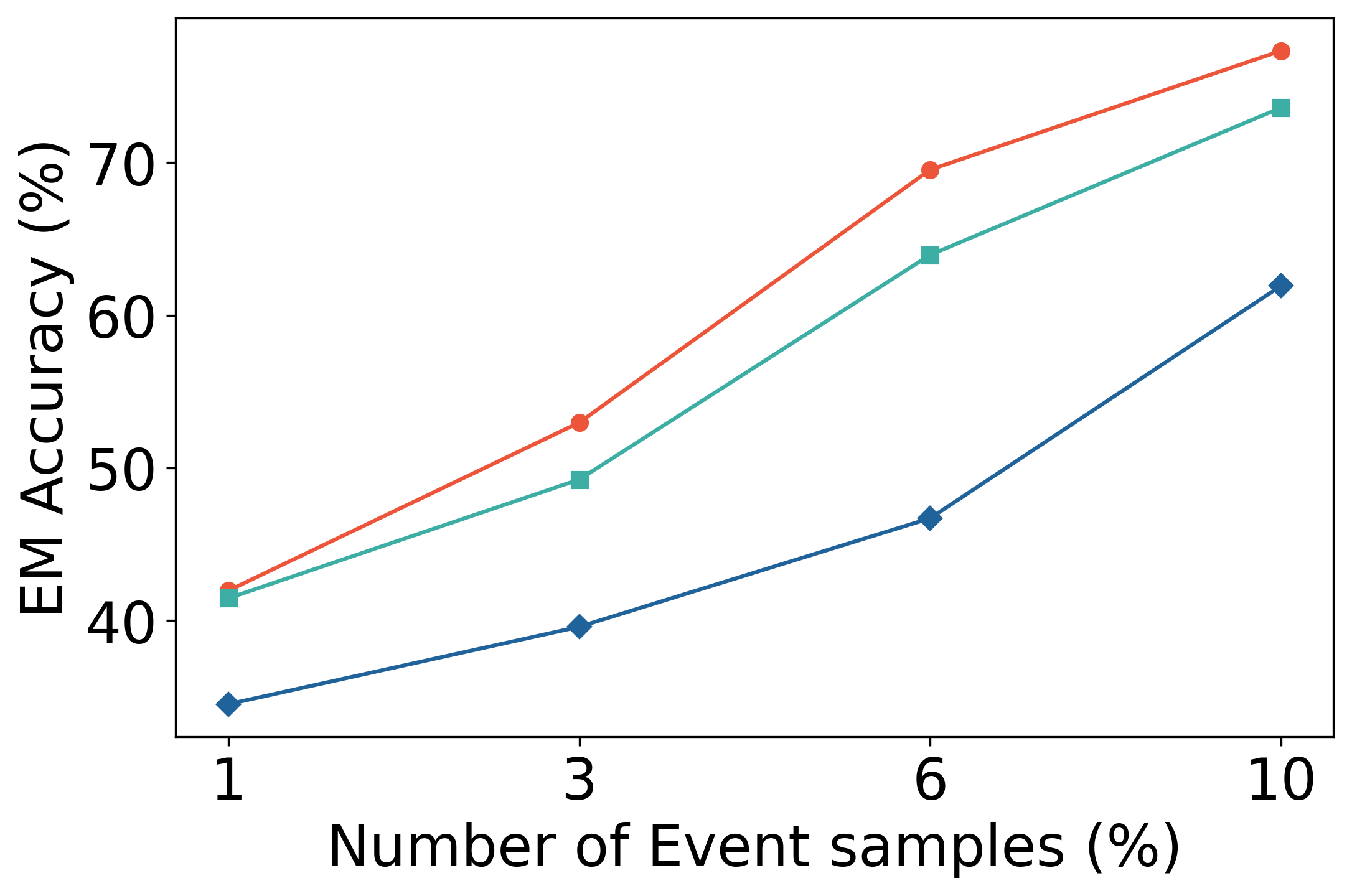}  
\end{minipage}
\begin{minipage}{.33\textwidth}
  \centering
  \includegraphics[width=\linewidth]{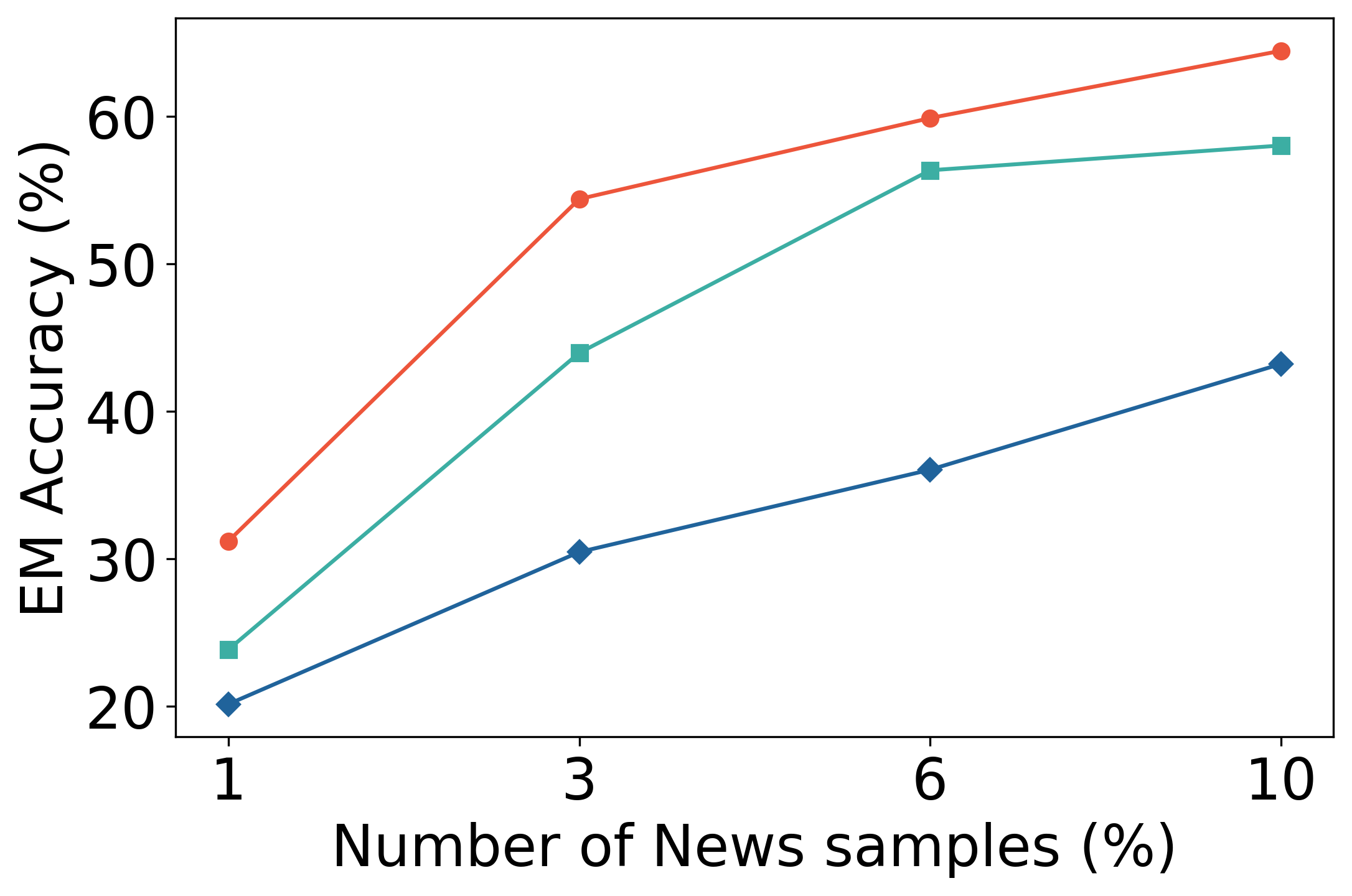}  
\end{minipage}
\begin{minipage}{.33\textwidth}
  \centering
  \includegraphics[width=\linewidth]{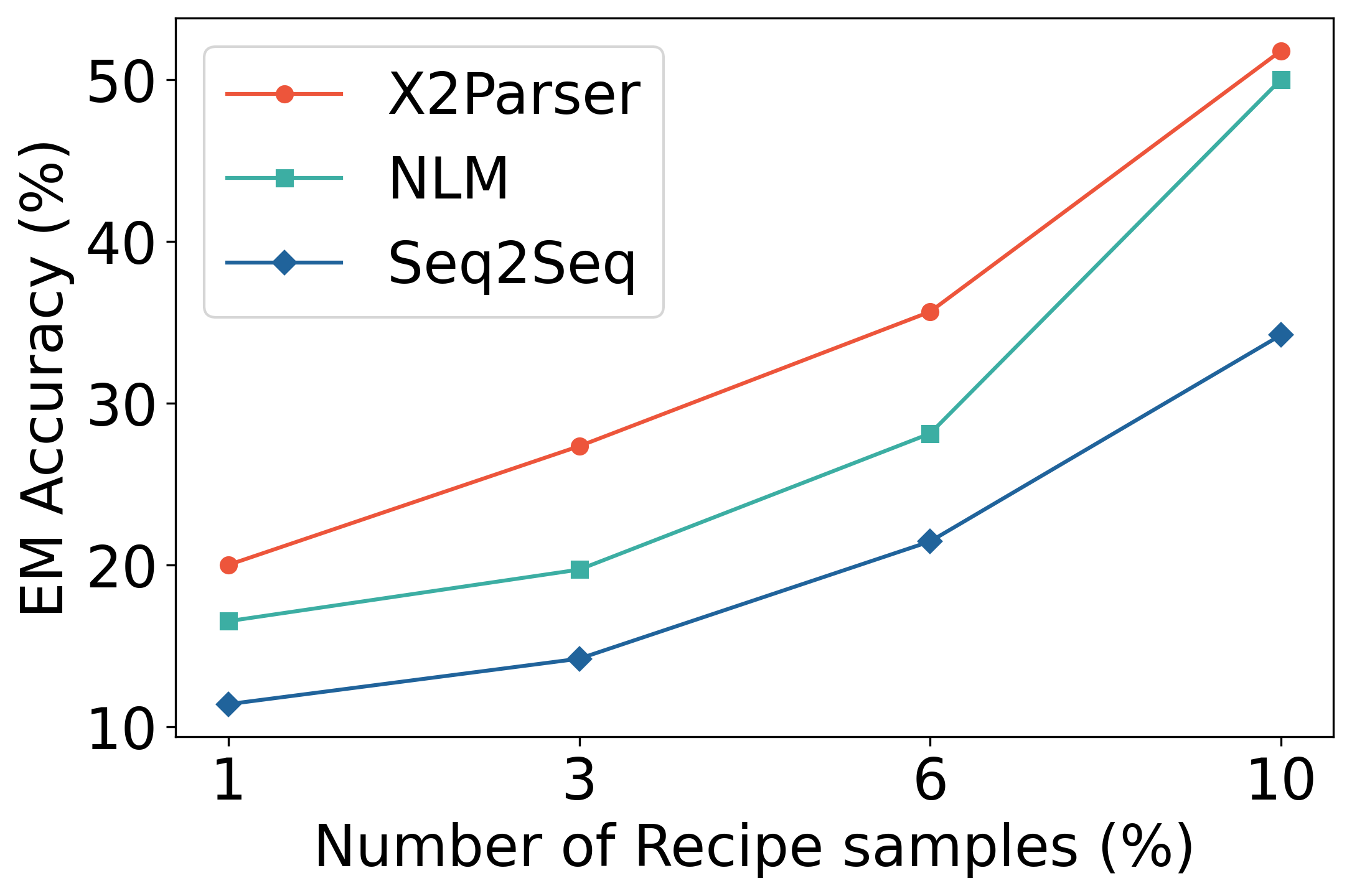}
\end{minipage}
\caption{Few-shot exact match results on the \textbf{cross-domain setting} for Event, News and Recipe target domains.}
\label{fig:results-cross-domain}
\end{figure*}

\begin{figure*}[!t]
\begin{minipage}{.33\textwidth}
  \centering
  \includegraphics[width=1\linewidth]{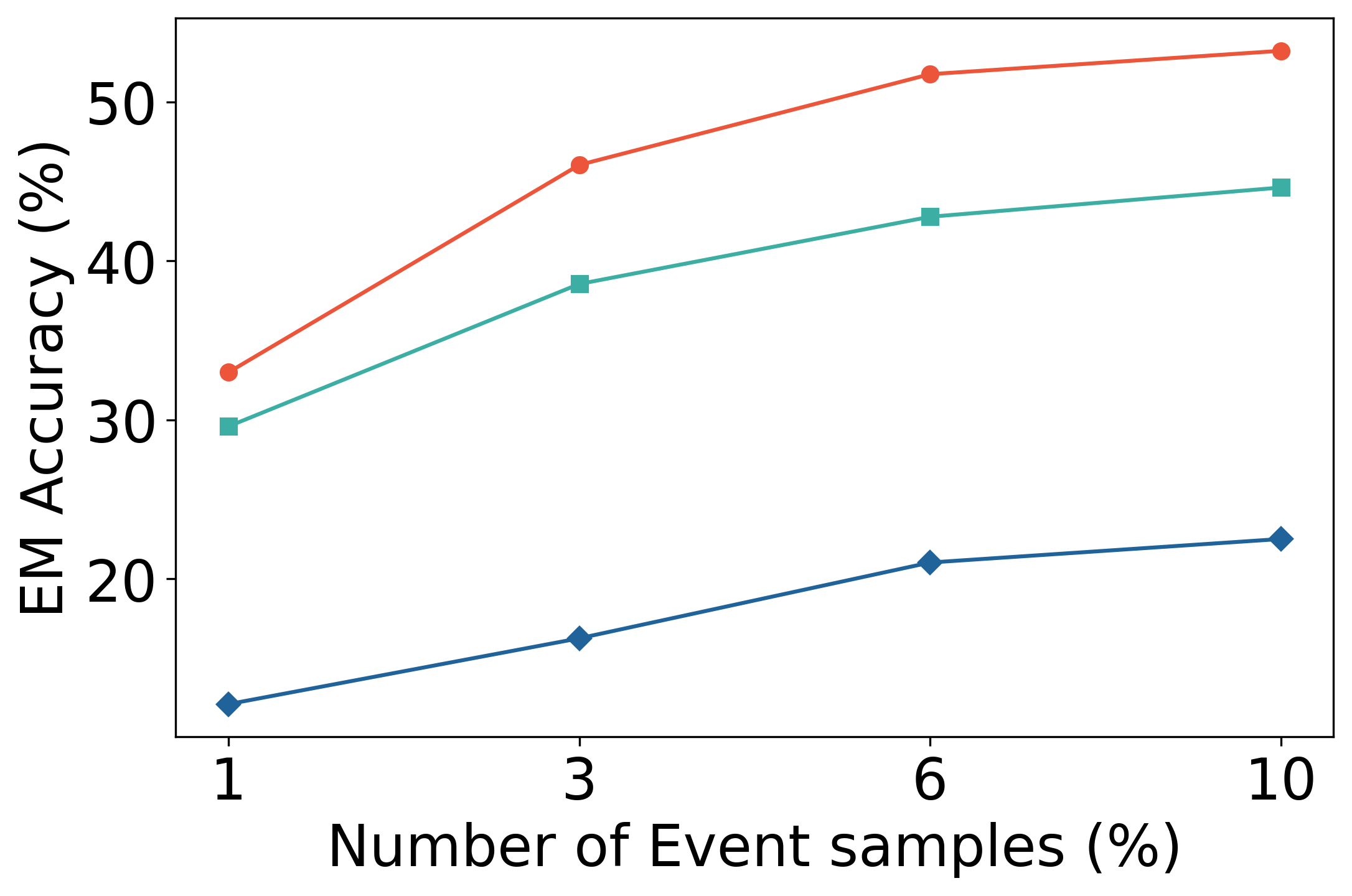}
\end{minipage}
\begin{minipage}{.33\textwidth}
  \centering
  \includegraphics[width=\linewidth]{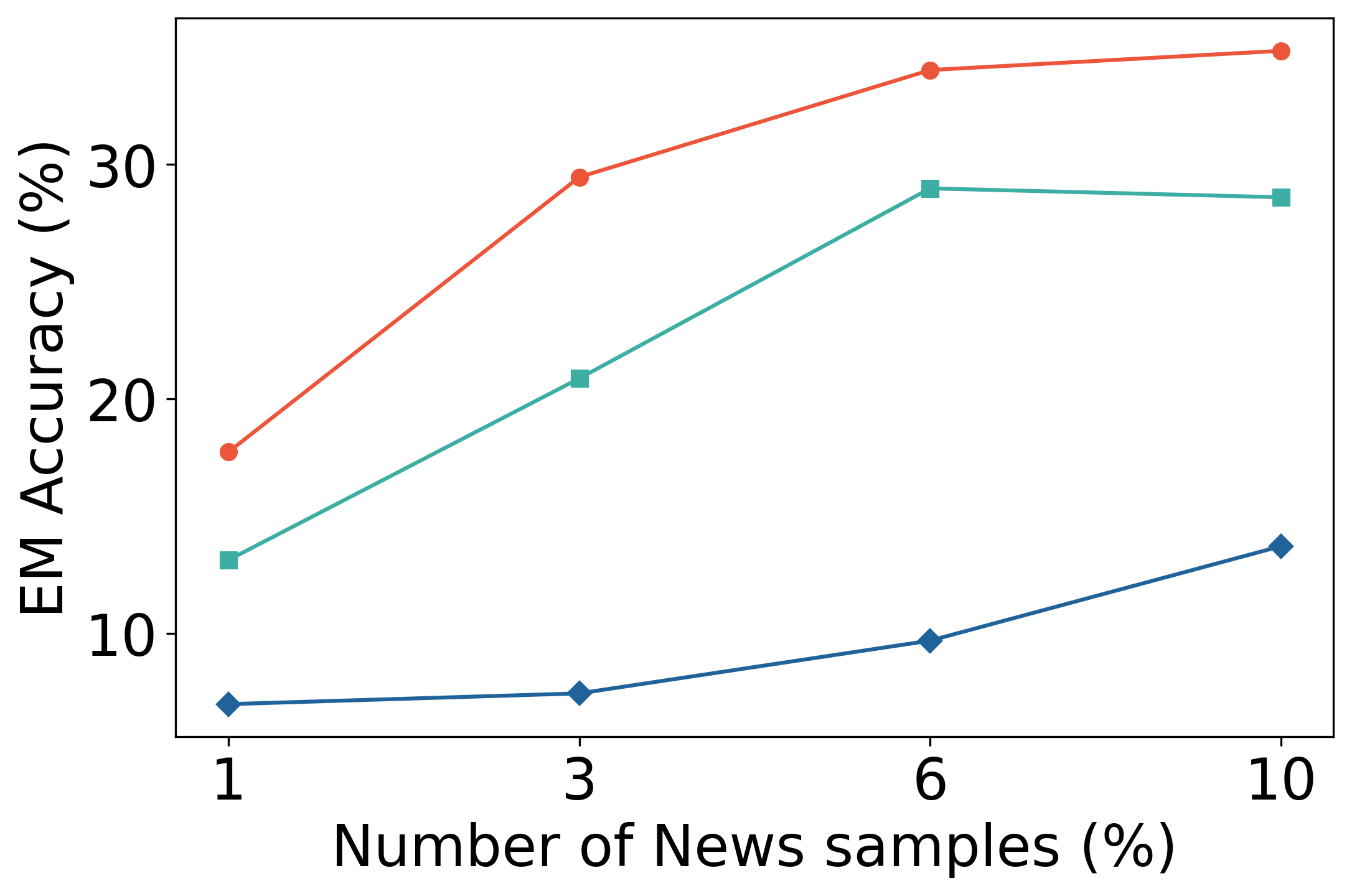}
\end{minipage}
\begin{minipage}{.33\textwidth}
  \centering
  \includegraphics[width=\linewidth]{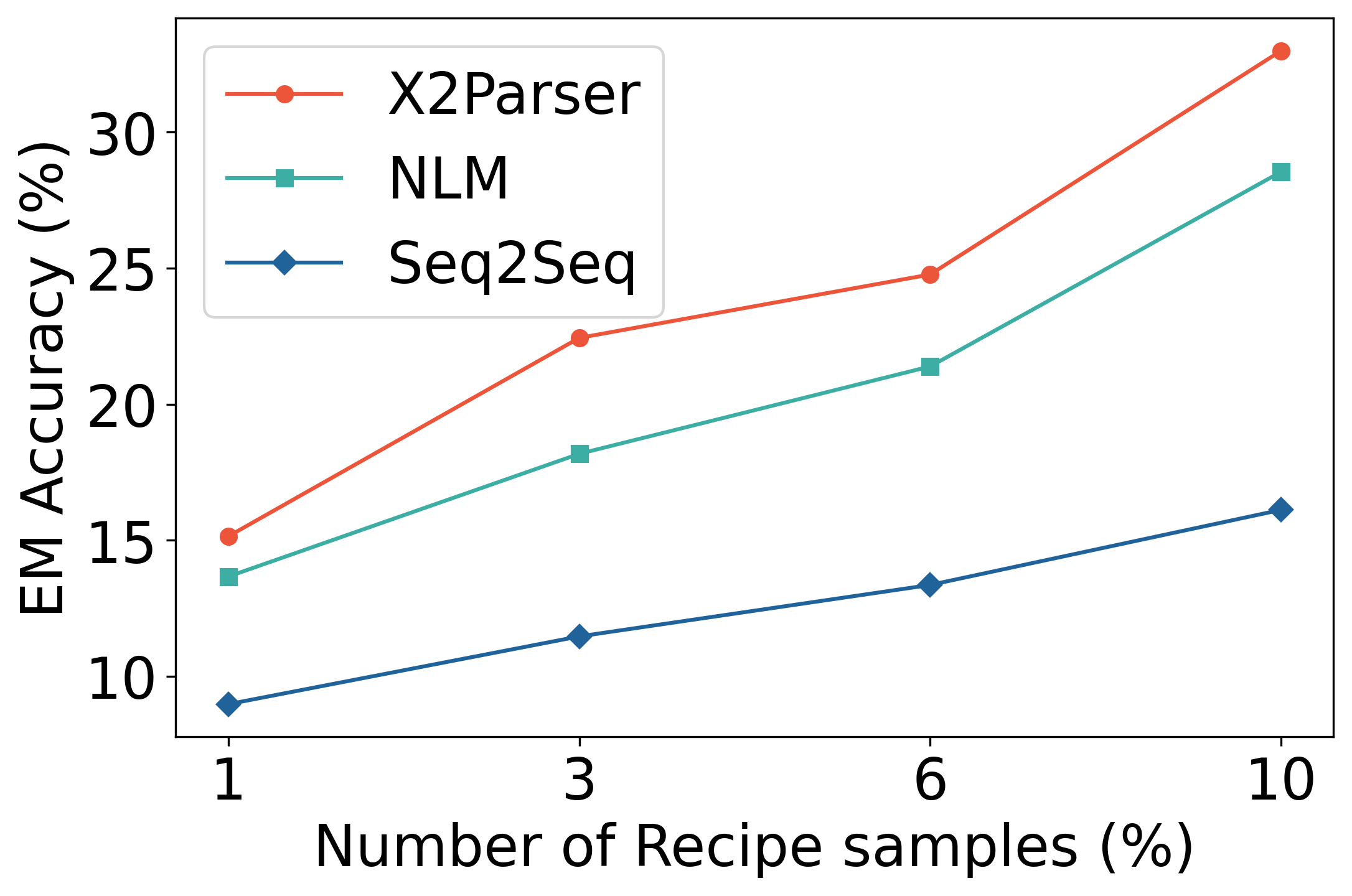}
\end{minipage}
\caption{Few-shot exact match results on the \textbf{cross-lingual cross-domain setting} for Event, News and Recipe target domains. The results are averaged over all target languages.}
\label{fig:results-x2}
\end{figure*}

\section{Results \& Discussion}
\subsection{Main Results}
\paragraph{Cross-Lingual Setting}
As we can see from Table~\ref{tab:cross_lingual_results}, X2Parser achieves similar performance in English compared to Seq2Seq-based models, while it significantly outperforms them in the zero-shot cross-lingual setting, with $\sim$10\% accuracy improvement on average. In the English training process, the Seq2Seq-based models can well learn the specific scope of tokens that need to be copied and assigned to a specific label type based on numerous training data.
However, these models will easily lose effectiveness when the input sequences are in target languages due to the inherent variances across languages and the difficulty of generating hierarchical representations. X2Parser separates the TCSP task into predicting intents and slots individually, which lowers the task difficulty and boosts its zero-shot adaptation ability to target languages.
Interestingly, we find that compared to \textit{Seq2Seq w/ XLM-R}, X2Parser greatly boosts the performance on target languages that are topologically close to English (e.g., French (fr)) with more than 10\% scores, while the improvements for languages that are topologically distant from English (e.g., Thai (th) and Hindi (hi)) are relatively limited. 
We argue that the large discrepancies between English and Thai make the representation alignment quality between English and Thai (Hindi) in XLM-R relatively low, and their different language patterns lead to unstable slot and intent predictions. These factors limit the improvement for X2Parser on the adaptation to topologically distant languages. 

From Table~\ref{tab:cross_lingual_results}, although NLM achieves marginally lower performance in English compared to \textit{Seq2Seq w/ XLM-R}, it produces significant improvements in target languages. This can be attributed to the fact that NLM leverages the same task decomposition as X2Parser, which further indicates the effectiveness of decomposing the TCSP task into intent and slot predictions for low-resource scenarios.
Additionally, X2Parser surpasses NLM by $\sim$2\% exact match accuracy on average in target languages. We conjecture that the stacked layers in NLM could make the model confused about which layer needs to generate which entity types, and this confusion is aggravated in the zero-shot cross-lingual setting where no training data are available. However, our fertility-based method helps the model implicitly learn the structure of hierarchical slots by predicting the number of labels for each token, which allows the slot classifier to predict the slot types more easily in the cross-lingual setting.

\paragraph{Cross-Domain Setting}
As shown in Table~\ref{tab:cross_domain_results}, X2Parser and NLM notably surpass the Seq2Seq model, with $\sim$15\% improvements on the averaged scores. This can be largely attributed to the effectiveness of our proposed task decomposition for low-resource scenarios. 
Seq2Seq models need to learn when to generate the label, when to copy tokens from the inputs, and when to produce the end of the label to generate hierarchical representations. This generation process requires a relatively large number of data samples to learn, which leads to the weak few-shot cross-domain performance for the Seq2Seq model.
Furthermore, X2Parser outperforms NLM, with a $\sim$2\% averaged score.
We conjecture that our fertility classifier guides the model to learn the inherent hierarchical information from the user queries, making it easier for the slot classifier to predict slot types for each token. However, the NLM's slot classifier, which consists of multiple stacked layers, needs to capture the hierarchical information and correctly assign slot labels of different levels to the corresponding stacked layer, which requires relatively larger data to learn.

\begin{table*}[!ht]
\renewcommand{\arraystretch}{1.3}
\centering
\begin{adjustbox}{width={0.999\textwidth},totalheight={\textheight},keepaspectratio}
\begin{tabular}{c|cccccccccc|cc}
\Xhline{2\arrayrulewidth}
\multirow{2}{*}{\textbf{Model}} & \multicolumn{2}{c}{\textbf{Spanish}} & \multicolumn{2}{c}{\textbf{French}} & \multicolumn{2}{c}{\textbf{German}} & \multicolumn{2}{c}{\textbf{Hindi}} & \multicolumn{2}{c|}{\textbf{Thai}} & \multicolumn{2}{c}{\textbf{Average}} \\
                       & \textbf{NN}           & \textbf{Nested}       & \textbf{NN}          & \textbf{Nested}       & \textbf{NN}          & \textbf{Nested}       & \textbf{NN}          & \textbf{Nested}      & \textbf{NN}         & \textbf{Nested}      & \textbf{NN}           & \textbf{Nested}       \\ \hline
Seq2Seq                & 56.21        & 29.38        & 48.11       & 32.83        & 46.02       & 20.25        & 37.84       & 22.30       & 33.27      & 13.56       & 44.29        & 23.66        \\
NLM                    & 65.65        & \textbf{41.95}        & 61.02       & 42.91        & 56.90       & 37.94        & 36.48       & 24.36       & 34.15      & 15.70       & 50.84        & 32.57        \\ \hline
X2Parser               & \textbf{66.69}        & 39.19        & \textbf{63.45}       & \textbf{44.28}        & \textbf{58.43}       & \textbf{39.71}        & \textbf{42.64}       & \textbf{28.55}       & \textbf{35.96}      & \textbf{16.67}       & \textbf{53.43}        & \textbf{33.68}        \\ \Xhline{2\arrayrulewidth}
\end{tabular}
\end{adjustbox}
\caption{Zero-shot cross-lingual exact match accuracies for nested and non-nested (NN) cases.}
\label{tab:nested_nonnested}
\end{table*}

\paragraph{Cross-Lingual Cross-Domain Setting}
From Table~\ref{tab:cross_lingual_cross_domain_results} and Figure~\ref{fig:results-combined}, we can further observe the effectiveness of our proposed task decomposition and X2Parser in the cross-lingual cross-domain setting. X2Parser and NLM consistently outperform the Seq2Seq model in all target languages of the target domains and boost the averaged exact match accuracy by $\sim$20\%. Additionally, from Table~\ref{tab:cross_lingual_cross_domain_results}, X2Parser also consistently outperforms NLM on all 11 domains and surpasses it by 3.84\% accuracy on average. From Figure~\ref{fig:results-combined}, X2Parser greatly improves on NLM in topologically distant languages (i.e., Hindi and Thai). It illustrates the powerful transferability and robustness of the fertility-based slot prediction that enables X2Parser to have a good zero-shot cross-lingual performance after it is fine-tuned to the target domain.

\subsection{Few-shot Analysis}
We conduct few-shot experiments using different sample sizes from the target domain for the cross-domain and cross-lingual cross-domain settings. The few-shot results on the Event, News, and Recipe target domains for both settings\footnote{We only report three domains due to the page limit, and place the full results for all 11 target domains in the Appendix~\ref{Appendix_C} and Appendix~\ref{Appendix_D}.} are shown in Figure~\ref{fig:results-cross-domain} and Figure~\ref{fig:results-x2}. We find that the performance of the Seq2Seq model is generally poor in both settings, especially when only 1\% of data samples are available.
With the help of the task decomposition, NLM and X2Parser remarkably outperform the Seq2Seq model in various target domains for both the cross-domain and cross-lingual cross-domain settings across different few-shot scenarios (from 1\% to 10\%).
Moreover, X2Parser consistently surpasses NLM for both the cross-domain and cross-lingual cross-domain settings in different few-shot scenarios, which further verifies the strong adaptation ability of our model. 

Interestingly, we observe that the improvement of X2Parser over Seq2Seq grows as the number of training samples increases. For example, in the cross-lingual cross-domain setting of the event domain, the improvement goes from 20\% to 30\% as the training data increases from 1\% to 10\%.
We hypothesize that in the low-resource scenario, the effectiveness of X2Parser will be greatly boosted when a relatively large number of data samples are available, while the Seq2Seq model needs much larger training data to achieve good performance.

\subsection{Analysis on Nested \& Non-Nested Data}
To further understand how our model improves the performance, we split the test data in the MTOP dataset~\cite{li2020mtop} into nested and non-nested samples. We consider the user utterances that do not have fine-grained intents and nested slots as the non-nested data sample and the rest of the data as the nested data sample. As we can see from Table~\ref{tab:nested_nonnested}, X2Parser significantly outperforms the Seq2Seq model on both nested and non-nested user queries with an average of $\sim$10\% accuracy improvement in both cases. In addition, X2Parser also consistently surpasses NLM on all target languages in both the nested and non-nested scenarios, except for the Spanish nested case, which further illustrates the stable and robust adaptation ability of X2Parser.

\begin{figure}[t!]
    \centering
    \resizebox{0.48\textwidth}{!}{  
    \includegraphics{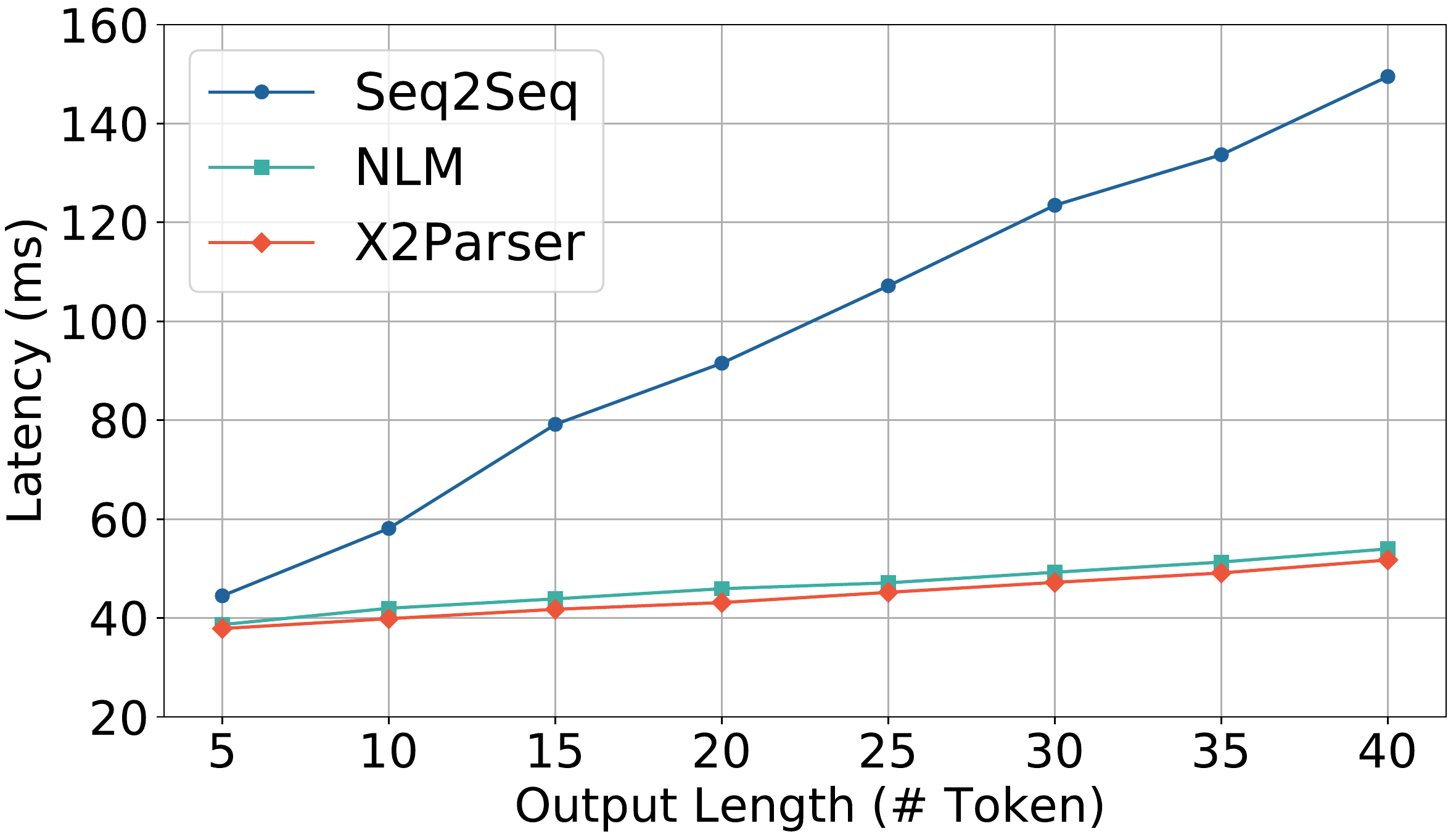}
    }
    \caption{Averaged latencies for our model and baselines on different output lengths of the MTOP dataset.}
    \label{fig:latency}
\end{figure}

\subsection{Latency Analysis}
We can see from Figure~\ref{fig:latency} that, as the output length increases, the latency discrepancy between the Seq2Seq-based model (Seq2Seq) and sequence labeling-based models (NLM and X2Parser) becomes larger, and when the output length reaches 40 tokens (around the maximum length in MTOP), X2Parser can achieve an up to 66\% reduction in latency compared to the Seq2Seq model. This can be attributed to the fact that the Seq2Seq model has to generate the outputs token by token, while X2Parser and NLM can directly generate all the outputs. In addition, the inference speed of X2Parser is slightly faster than that of NLM. This is because NLM uses several stacked layers to predict slot entities of different levels, and the higher-level layer has to wait for the predictions from the lower-level layer, which slightly decreases the inference speed.

\section{Conclusion}
In this paper, we develop a transferable and non-autoregressive model (X2Parser) for the TCSP task that can better adapt to target languages and domains with a faster inference speed. Unlike previous TCSP models that learn to generate hierarchical representations, we propose to decompose the task into intent and slot predictions so as to lower the difficulty of the task, and then we cast both prediction tasks into sequence labeling problems. After that, we further propose a fertility-based method to cope with the slot prediction task where each token could have multiple labels. Results illustrate that X2Parser significantly outperforms strong baselines in all low-resource settings. Furthermore, our model is able to reduce the latency by up to 66\% compared to the generation-based model.

\section*{Acknowledgement}
We want to say thanks to the anonymous reviewers for the insightful reviews and constructive feedback. This work is partially funded by ITF/319/16FP and MRP/055/18 of the Innovation Technology Commission, the Hong Kong SAR Government.

\bibliographystyle{acl_natbib}
\bibliography{acl2021}

\clearpage
\appendix
\section{Intent Label Construction}
\label{Appendix_A}
In this section, we further describe how we convert the fine-grained intent prediction into a sequence labeling task (each token has only one label). We use a few examples to illustrate our intent label construction method.

\begin{figure}[!ht]
    \centering
    \resizebox{0.49\textwidth}{!}{  
    \includegraphics{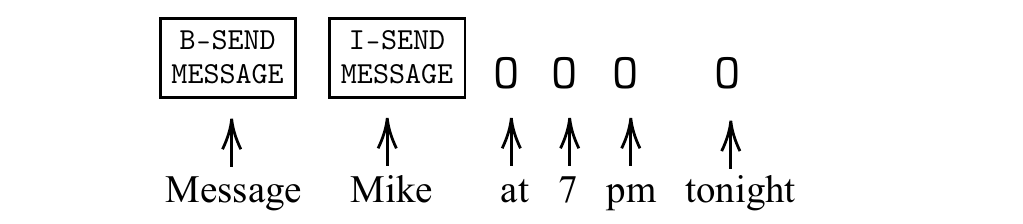}
    }
    \caption{A labeling example for non-nested intent.}
    \label{fig:nonnested_intent}
\end{figure}

As illustrated in Figure~\ref{fig:nonnested_intent}, when there are no nested intents in the input utterance, we follow the BIO structure to give intent labels. 

\begin{figure}[!ht]
    \centering
    \resizebox{0.49\textwidth}{!}{  
    \includegraphics{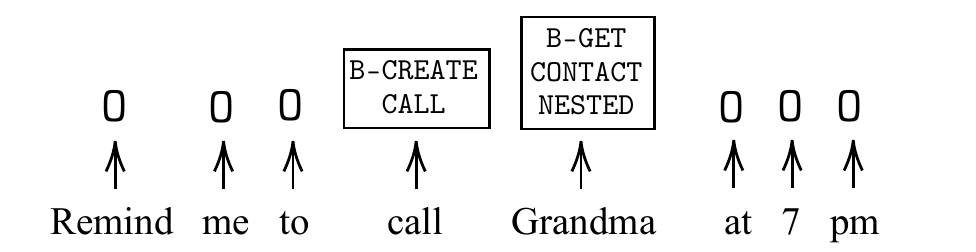}
    }
    \caption{A labeling example for nested intent.}
    \label{fig:nested_intent}
\end{figure}

We can see from Figure~\ref{fig:nested_intent} that ``call Grandma'' is a \texttt{CREATE-CALL} intent and ``Grandma'' is a \texttt{GET-CONTACT} intent. Hence, the \texttt{GET-CONTACT} intent is nested in the \texttt{CREATE-CALL} intent. We use a special intent label (with ``NESTED'') for the ``\texttt{GET-CONTACT}'' intent (\texttt{B-GET-CONTACT-NESTED}) to represent that this intent is nested in another intent, and hence, the scope of the \texttt{CREATE-CALL} intent is automatically expanded from ``call'' to ``call Grandma''.~\footnote{We notice that if two intents have overlaps, one intent either fully covers the other intent or is fully covered by the other intent.}

Note that we cannot apply this labeling method to the slot prediction since one token in the user utterance could be the starting token for more than one slot entity. If that is the case, we have to use more than one slot label for this token to denote the starting position for each slot entity. 
Given that in the MTOP dataset, one token will not be the starting token of more than one intent, we can apply this method for the intent label construction.
In the future, when more complex and sophisticated datasets are collected for the task-oriented compositional semantic parsing task, where there could exist more than one intent label for each token, we can always use the fertility-based method (currently applied for the slot prediction) for the intent prediction.

\section{Data Statistics}
The data statistics for MTOP are shown in Table~\ref{tab:data_statistics}.
\label{Appendix_B}

\begin{table*}[!ht]
\renewcommand{\arraystretch}{1.05}
\centering
\begin{adjustbox}{width={0.85\textwidth},totalheight={\textheight},keepaspectratio}
\begin{tabular}{p{3cm}|cccccc|cc}
\Xhline{2\arrayrulewidth}
\multirow{2}{*}{\textbf{Domain}} & \multicolumn{6}{c|}{\textbf{Number of Utterances}}              & \multirow{2}{*}{\begin{tabular}[c]{@{}c@{}}\textbf{Intent}\\ \textbf{Types}\end{tabular}} & \multirow{2}{*}{\begin{tabular}[c]{@{}c@{}}\textbf{Slot}\\ \textbf{Types}\end{tabular}} \\ \cline{2-7}
& \textbf{English} & \textbf{German} & \textbf{French} & \textbf{Spanish} & \textbf{Hindi}  & \textbf{Thai}   & & \\ \hline
\textbf{Alarm}                   & 1,783   & 1,581  & 1,706  & 1,377   & 1,510  & 1,783  & 6                                                                       & 5                                                                     \\
\textbf{Calling}                 & 2,872   & 2,797  & 2,057  & 2,515   & 2,490  & 2,872  & 19                                                                      & 14                                                                    \\
\textbf{Event}                   & 1,081   & 1,051  & 1,115  & 911     & 988    & 1,081  & 12                                                                      & 12                                                                    \\
\textbf{Messaging}               & 1,053   & 1,239  & 1,335  & 1,164   & 1,082  & 1,053  & 7                                                                       & 15                                                                    \\
\textbf{Music}                   & 1,648   & 1,499  & 1,312  & 1,509   & 1,418  & 1,648  & 27                                                                      & 12                                                                    \\
\textbf{News}                    & 1,393   & 905    & 1,052  & 1,130   & 930    & 1,393  & 3                                                                       & 6                                                                     \\
\textbf{People}                  & 1,449   & 1,392  & 763    & 1,408   & 1,168  & 1,449  & 17                                                                      & 16                                                                    \\
\textbf{Recipes}                 & 1,586   & 1,002  & 762    & 1,382   & 929    & 1,586  & 3                                                                       & 18                                                                    \\
\textbf{Reminder}                & 2,439   & 2,321  & 2,202  & 1,811   & 1,833  & 2,439  & 19                                                                      & 17                                                                    \\
\textbf{Timer}                   & 1,358   & 1,014  & 1,165  & 1,159   & 1,047  & 1,358  & 9                                                                       & 5                                                                     \\
\textbf{Weather}                 & 2,126   & 1,785  & 1,990  & 1,816   & 1,800  & 2,126  & 4                                                                       & 4                                                                     \\ \hline
\textbf{Total}                   & 18,788  & 16,585 & 15,459 & 16,182  & 15,195 & 18,788 & 117                                                                     & 78                                                                    \\ \Xhline{2\arrayrulewidth}
\end{tabular}
\end{adjustbox}
\caption{Data statistics of the MTOP dataset. The data are roughly divided into a 70:10:20 percent split for train, eval and test}
\label{tab:data_statistics}
\end{table*}

\section{Few-shot Cross-Domain Results}
\begin{figure*}[!ht]
\begin{minipage}{.33\textwidth}
  \centering
  \includegraphics[width=1\linewidth]{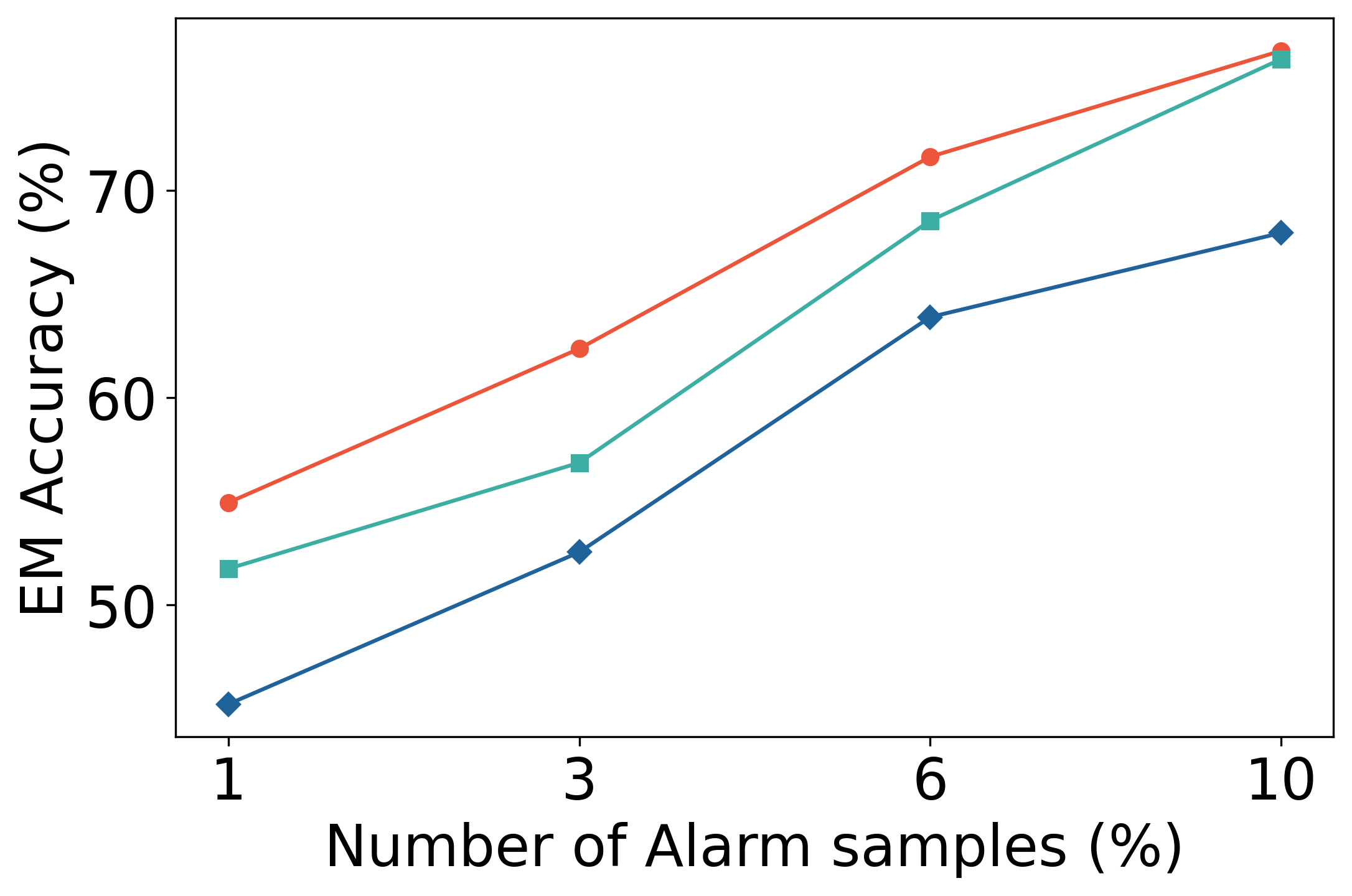}  
\end{minipage}
\begin{minipage}{.33\textwidth}
  \centering
  \includegraphics[width=\linewidth]{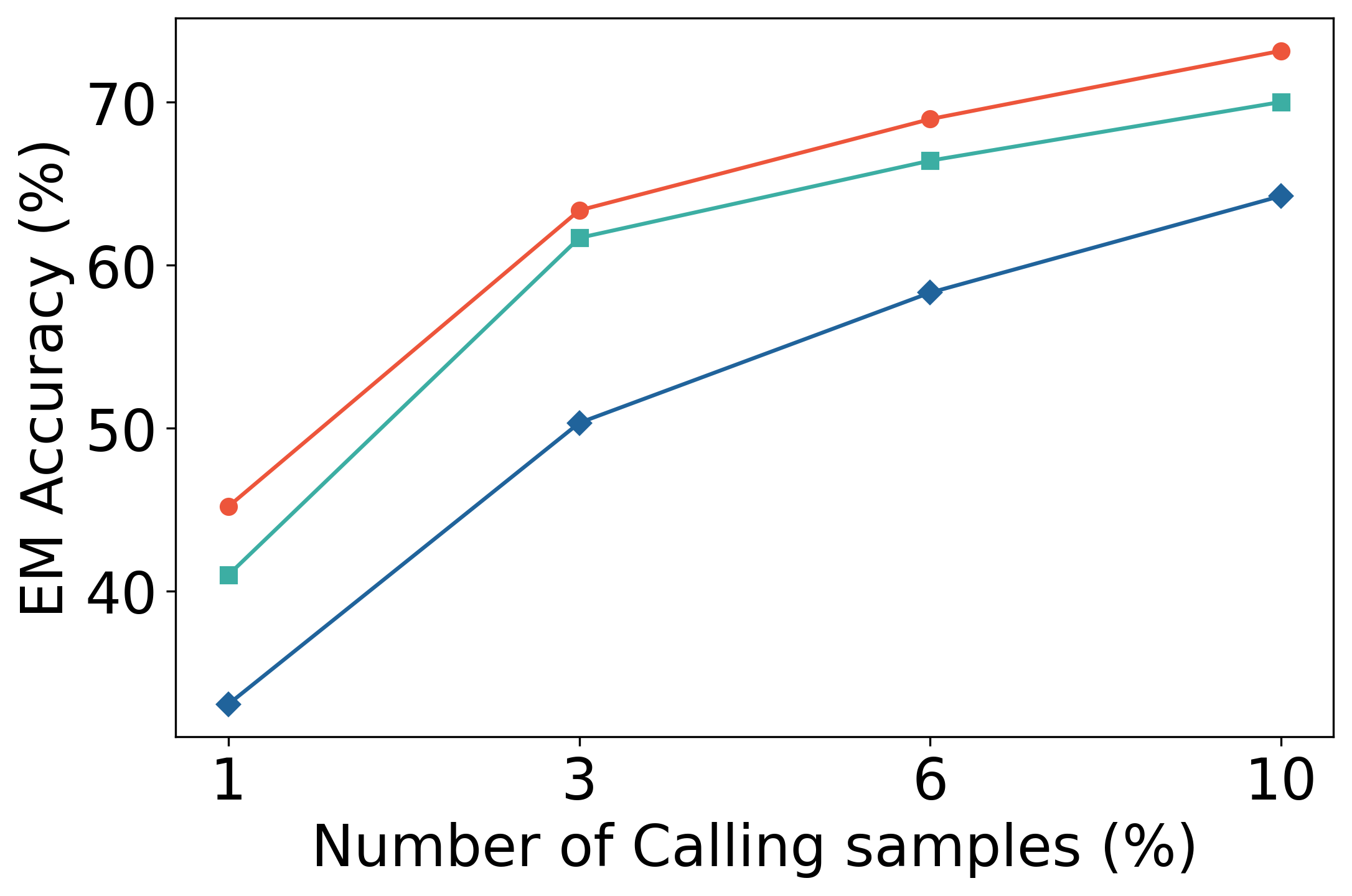}  
\end{minipage}
\begin{minipage}{.33\textwidth}
  \centering
  \includegraphics[width=\linewidth]{img/cross_domain/Event.png}
\end{minipage}
\begin{minipage}{.33\textwidth}
  \centering
  \includegraphics[width=1\linewidth]{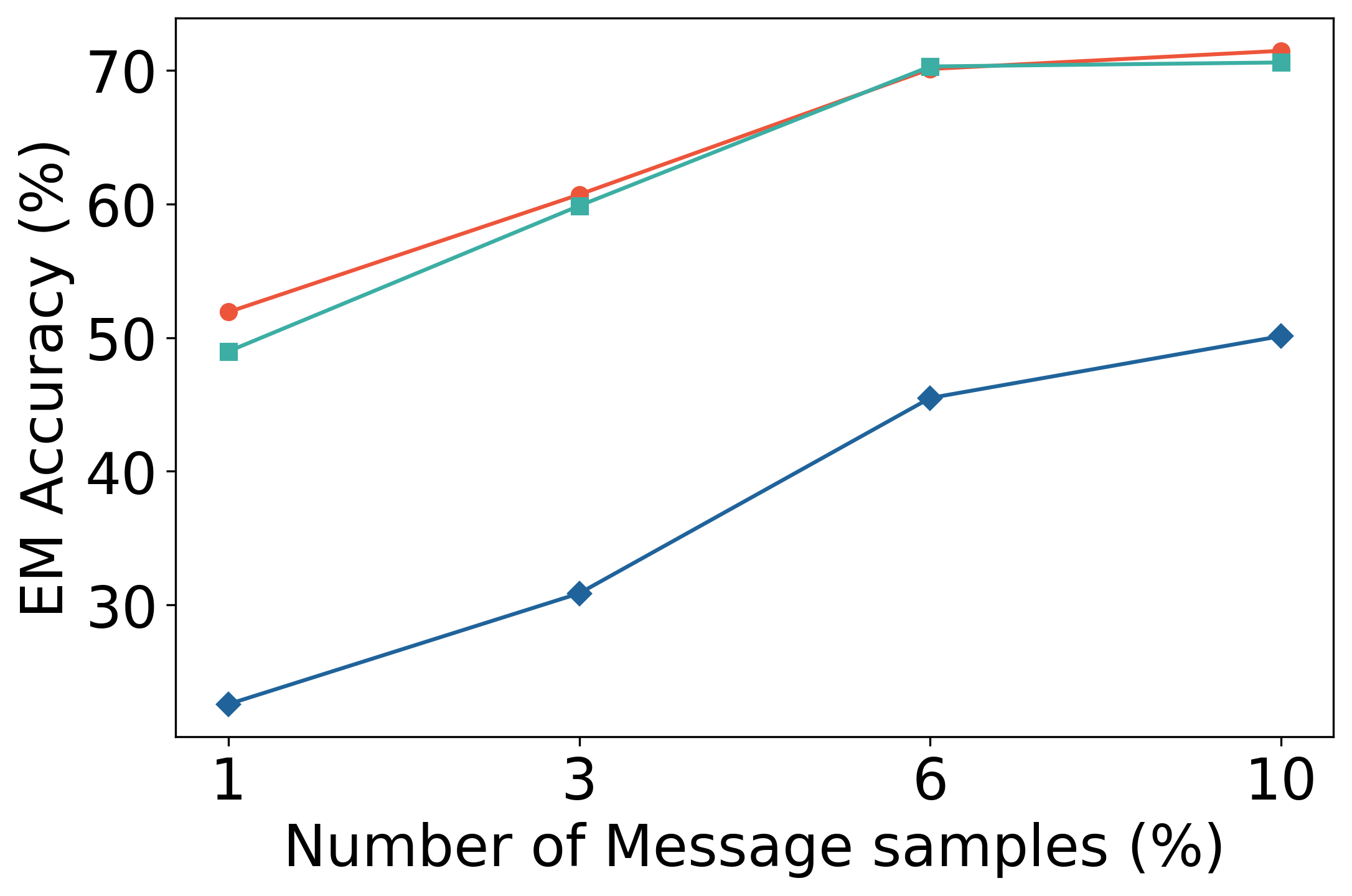}  
\end{minipage}
\begin{minipage}{.33\textwidth}
  \centering
  \includegraphics[width=\linewidth]{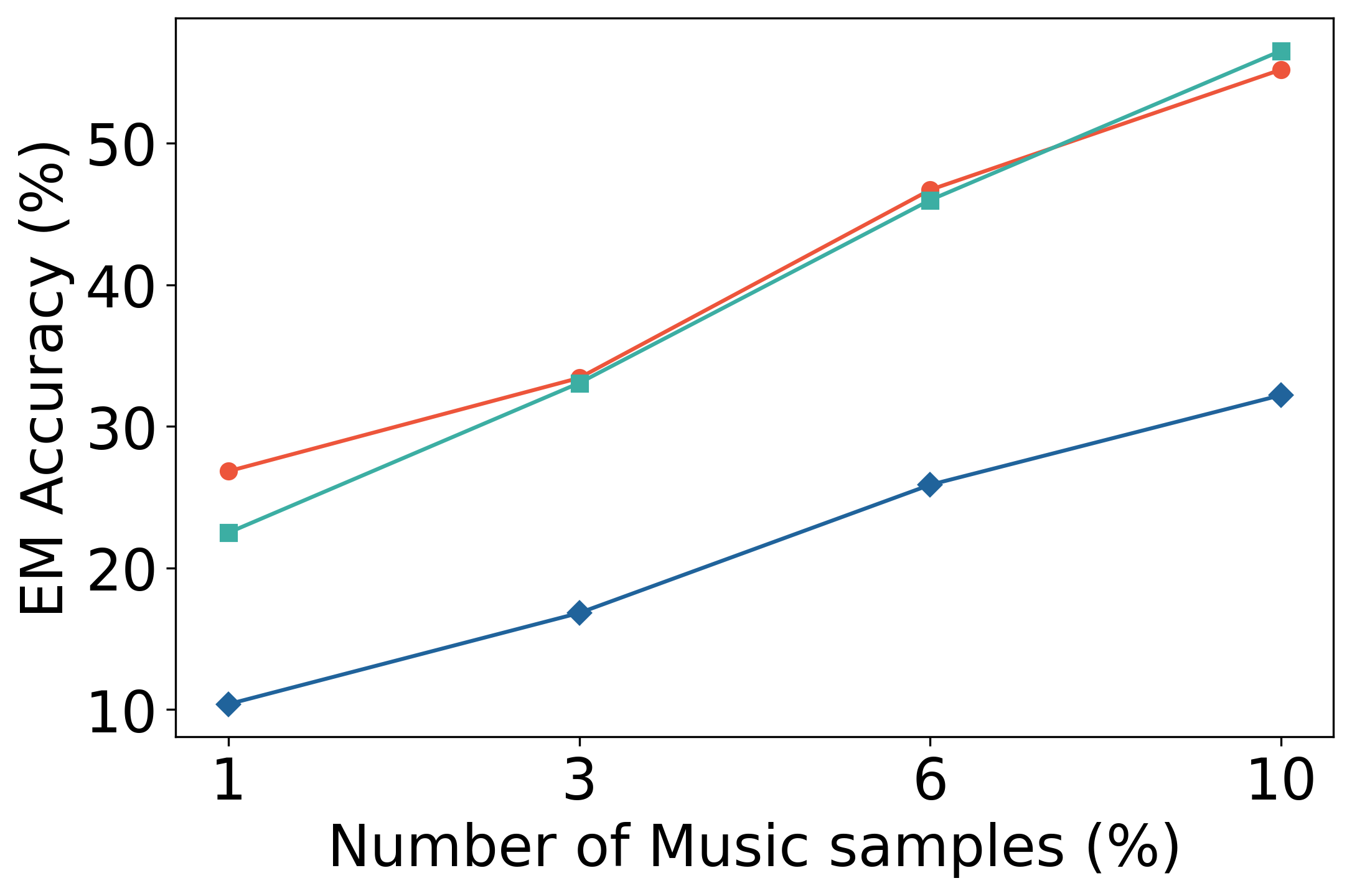}  
\end{minipage}
\begin{minipage}{.33\textwidth}
  \centering
  \includegraphics[width=\linewidth]{img/cross_domain/News.png}
\end{minipage}
\begin{minipage}{.33\textwidth}
  \centering
  \includegraphics[width=1\linewidth]{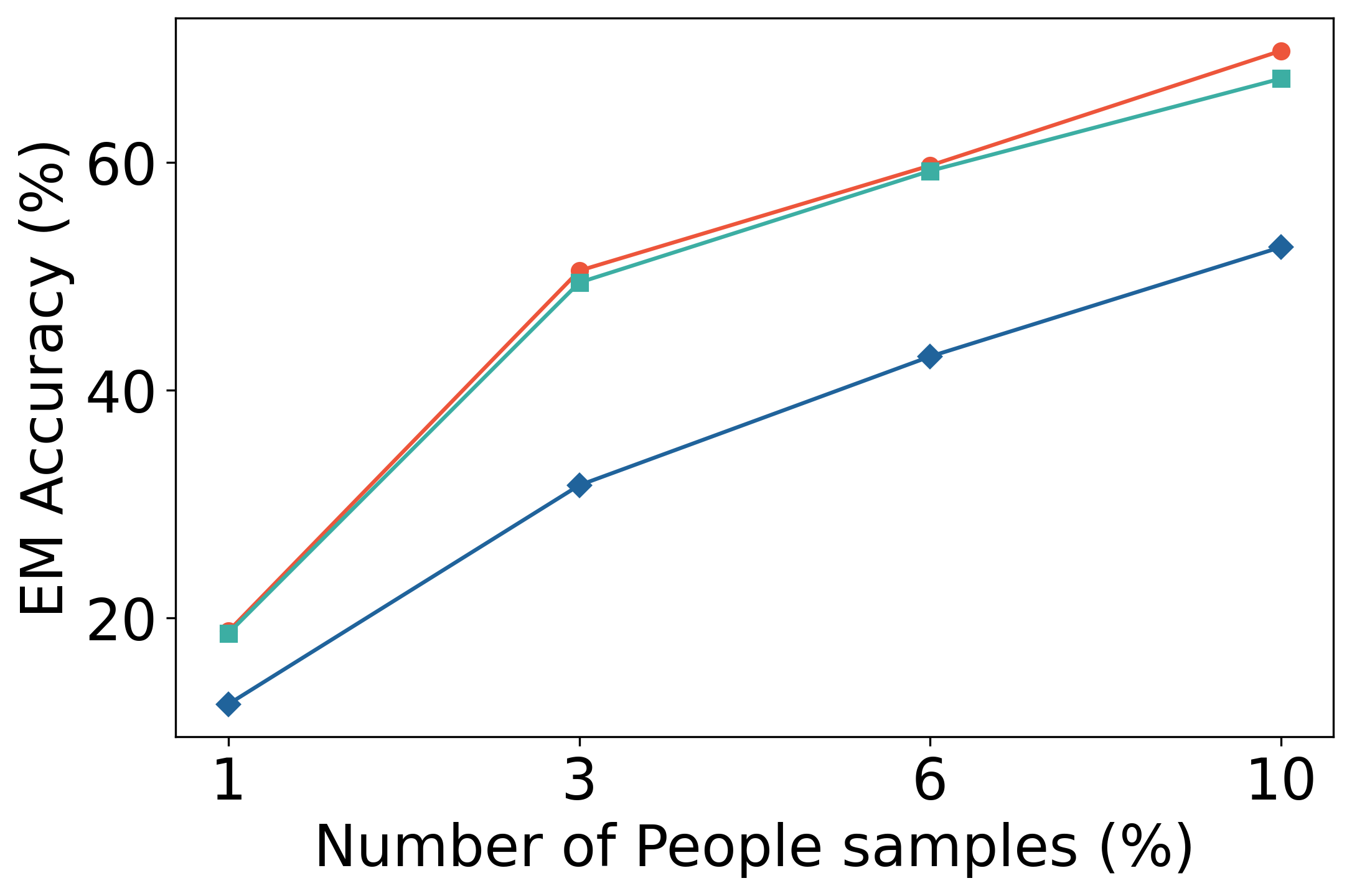}  
\end{minipage}
\begin{minipage}{.33\textwidth}
  \centering
  \includegraphics[width=\linewidth]{img/cross_domain/Recipe.png}  
\end{minipage}
\begin{minipage}{.33\textwidth}
  \centering
  \includegraphics[width=\linewidth]{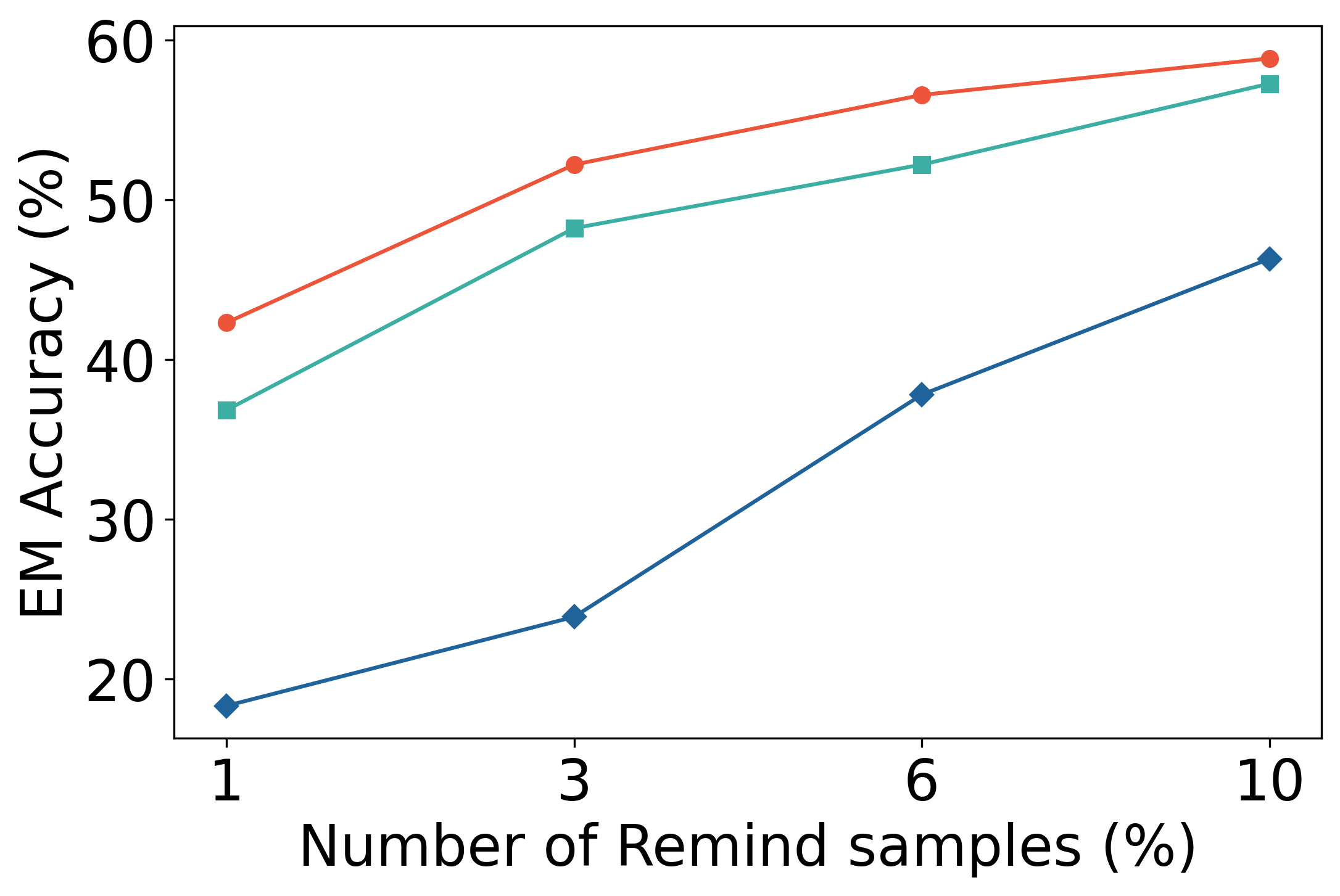}
\end{minipage}
\begin{minipage}{.33\textwidth}
  \centering
  \includegraphics[width=1\linewidth]{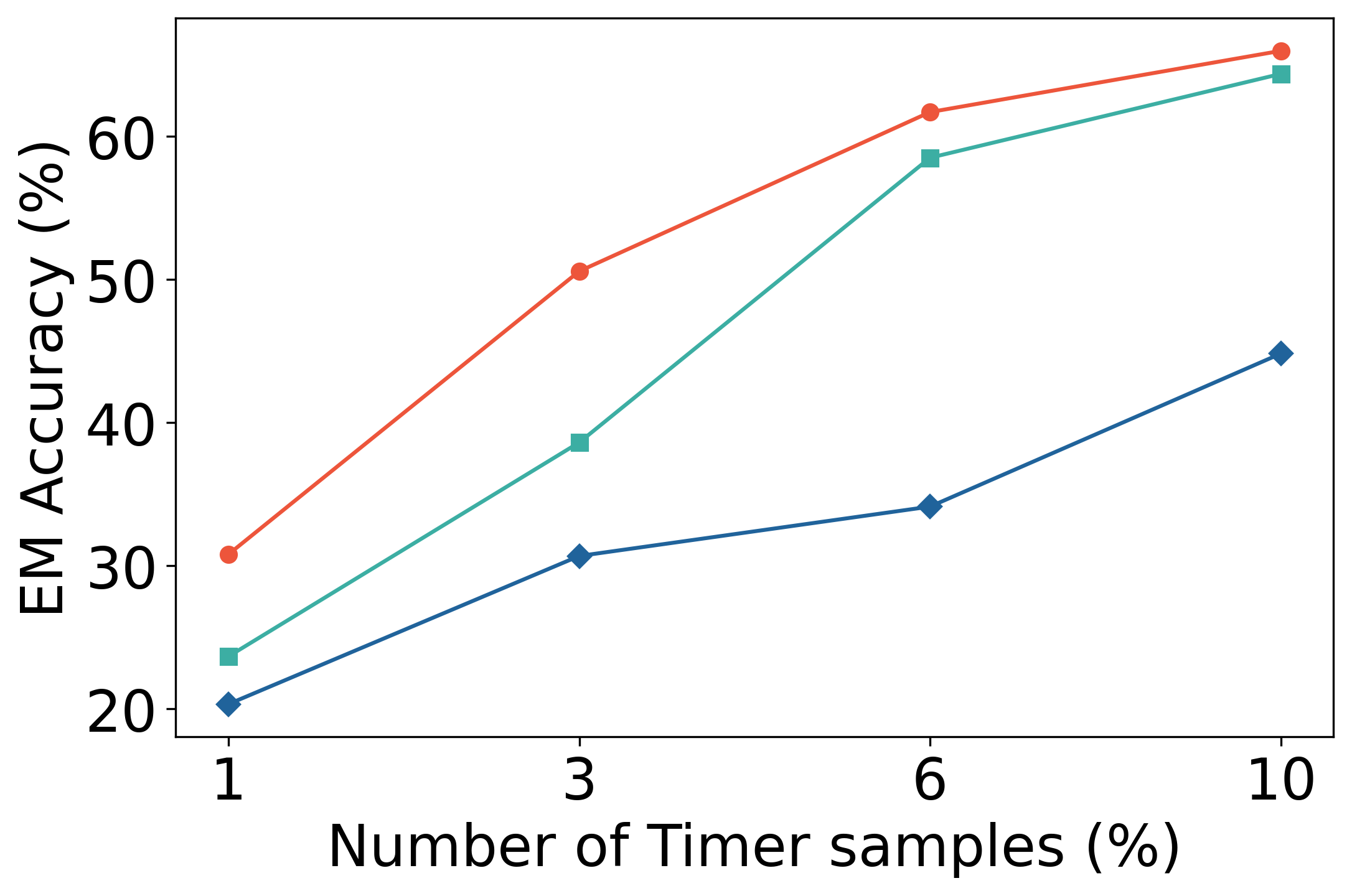}  
\end{minipage}
\begin{minipage}{.33\textwidth}
  \centering
  \includegraphics[width=\linewidth]{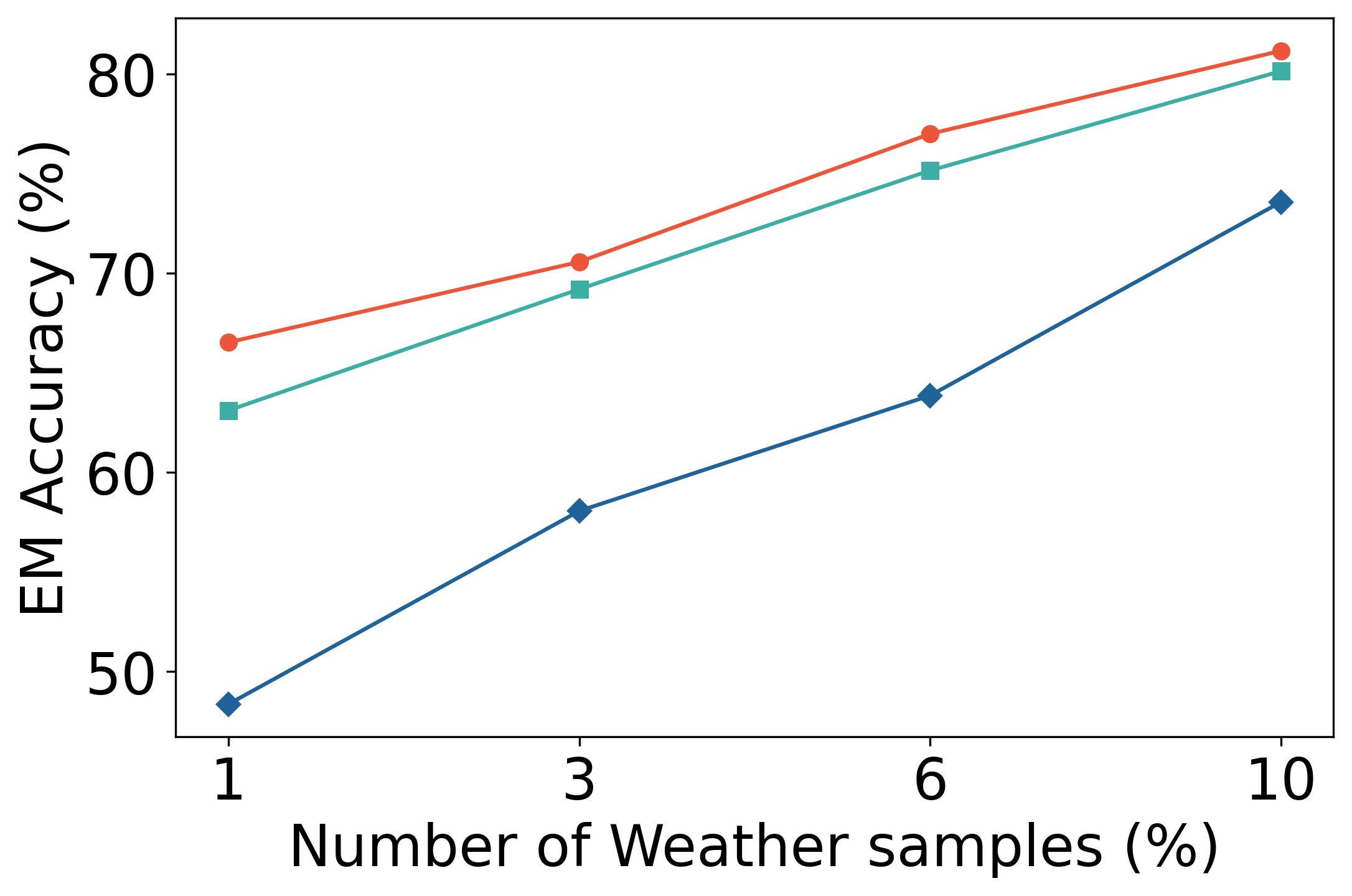}  
\end{minipage}
\caption{Few-shot exact match accuracies for the \textbf{cross-domain setting} across all 11 target domains.}
\label{fig:results-all-cross-domain}
\end{figure*}

Full few-shot cross-domain results across all 11 target domains are shown in Figure~\ref{fig:results-all-cross-domain} and Table~\ref{tab:results-complete-cross-domain}. \label{Appendix_C}

\section{Few-shot Cross-Lingual Cross-Domain Results}
\label{Appendix_D}

\begin{figure*}[!ht]
\begin{minipage}{.33\textwidth}
  \centering
  \includegraphics[width=1\linewidth]{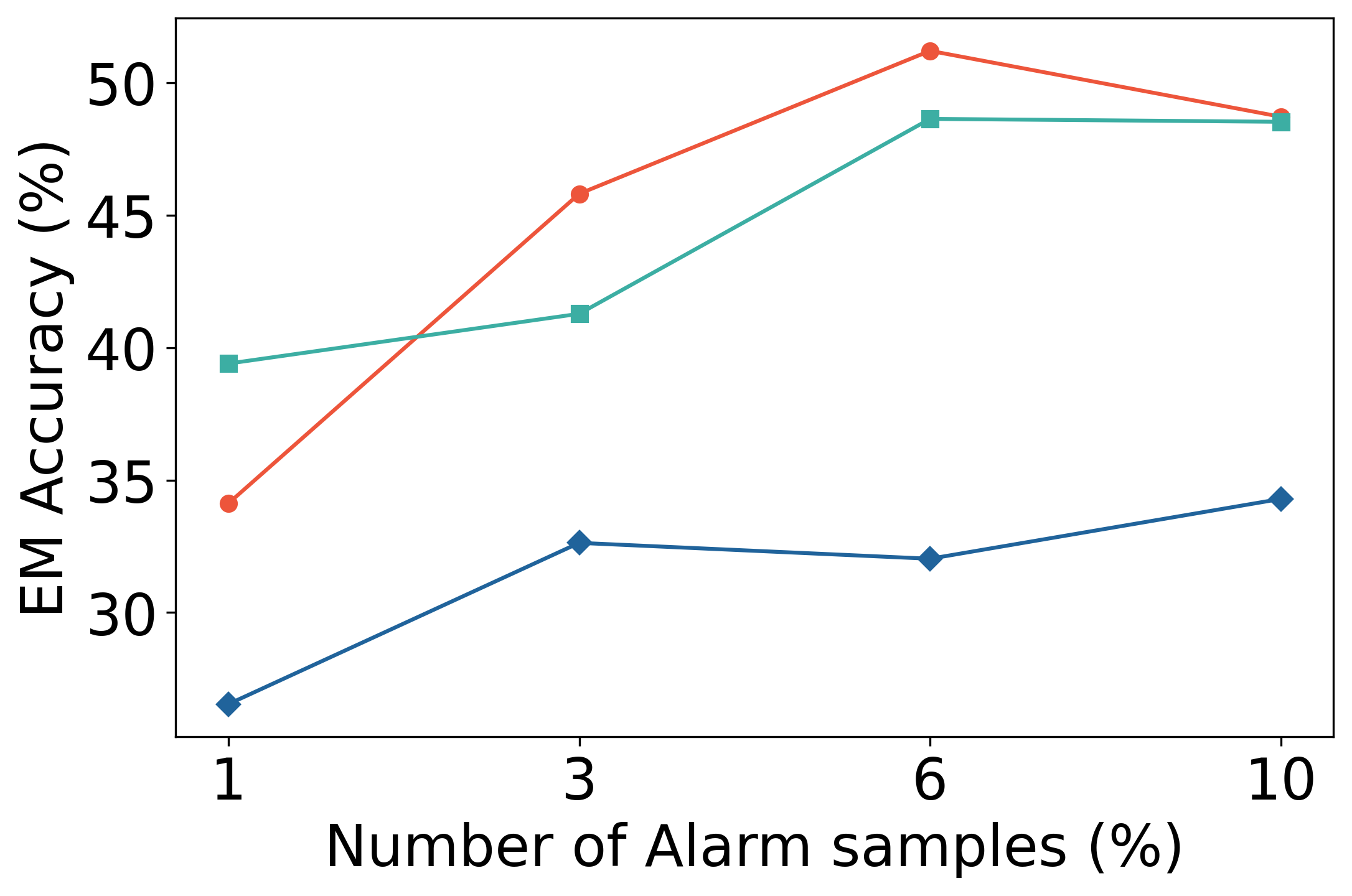}  
\end{minipage}
\begin{minipage}{.33\textwidth}
  \centering
  \includegraphics[width=\linewidth]{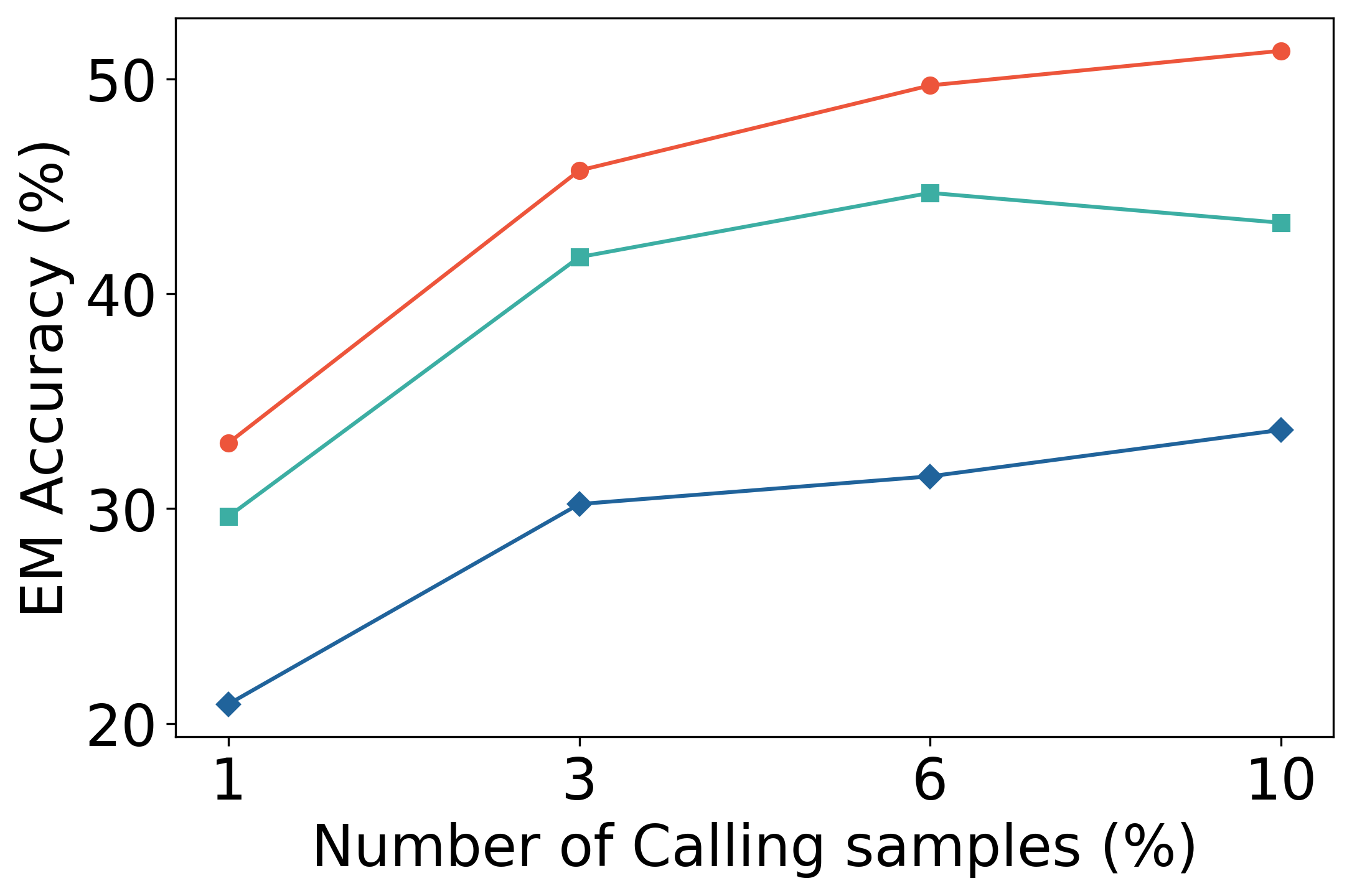}  
\end{minipage}
\begin{minipage}{.33\textwidth}
  \centering
  \includegraphics[width=\linewidth]{img/X2/Event.png}
\end{minipage}
\begin{minipage}{.33\textwidth}
  \centering
  \includegraphics[width=1\linewidth]{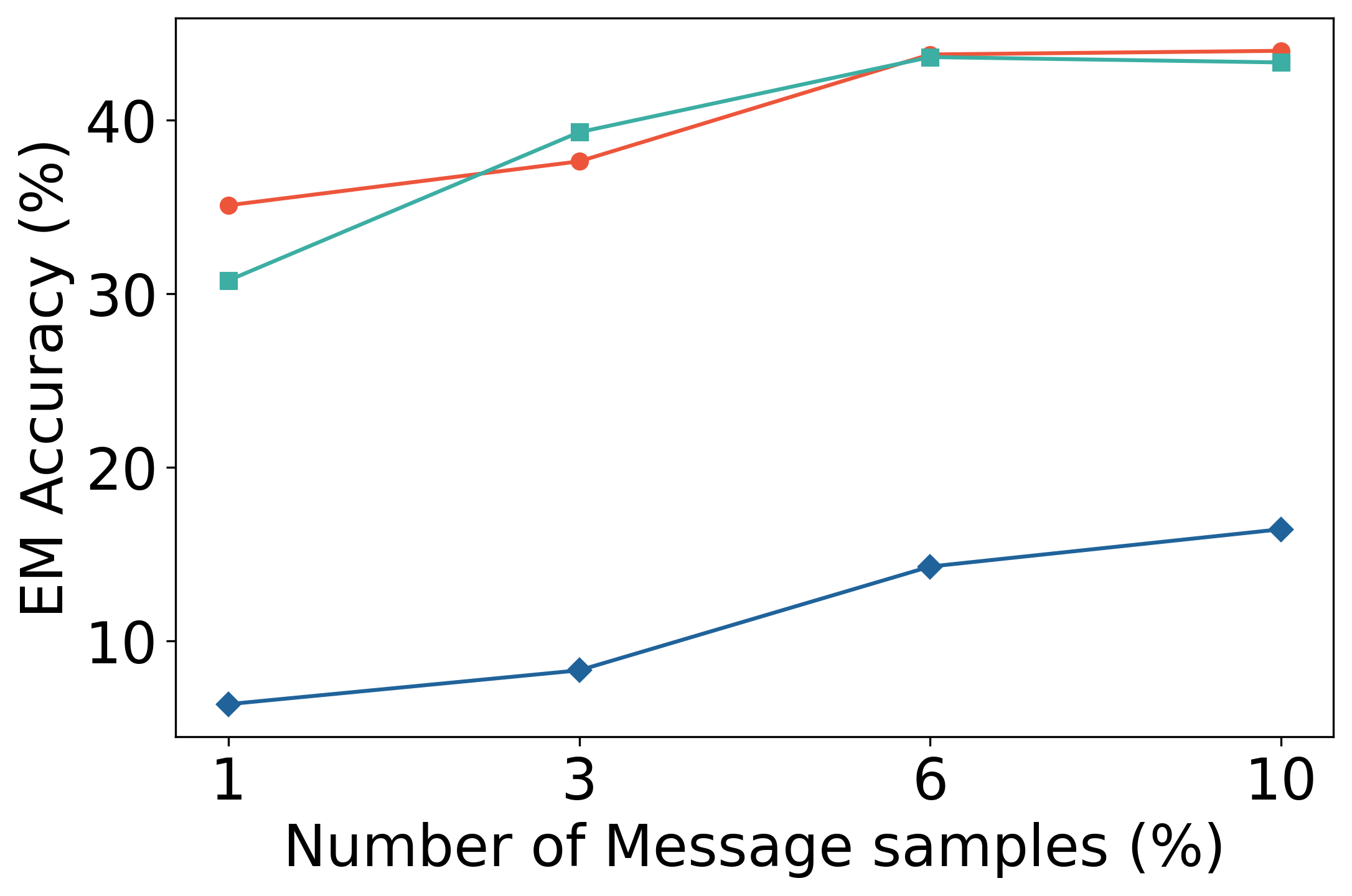}  
\end{minipage}
\begin{minipage}{.33\textwidth}
  \centering
  \includegraphics[width=\linewidth]{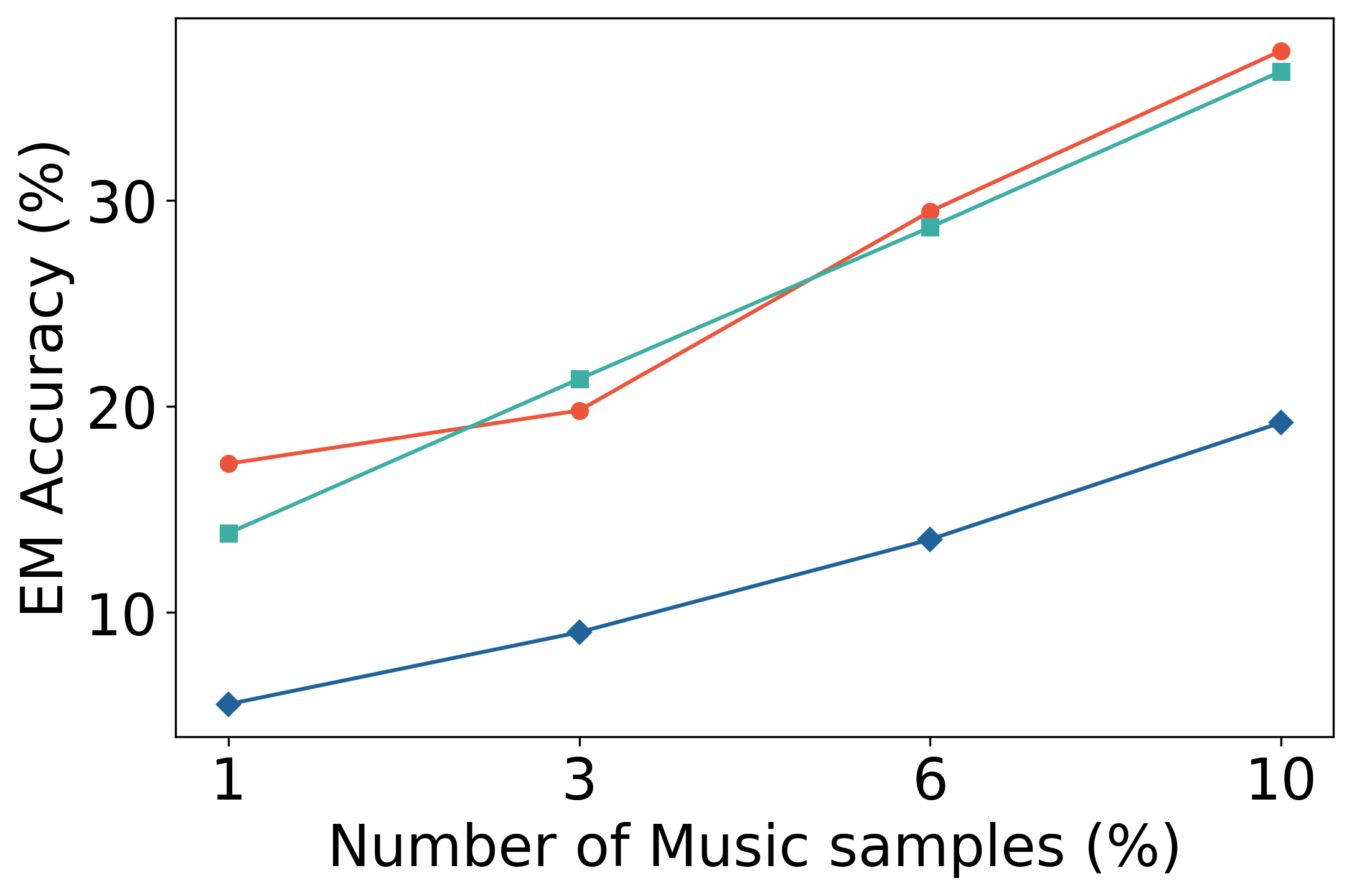}  
\end{minipage}
\begin{minipage}{.33\textwidth}
  \centering
  \includegraphics[width=\linewidth]{img/X2/News.png}
\end{minipage}
\begin{minipage}{.33\textwidth}
  \centering
  \includegraphics[width=1\linewidth]{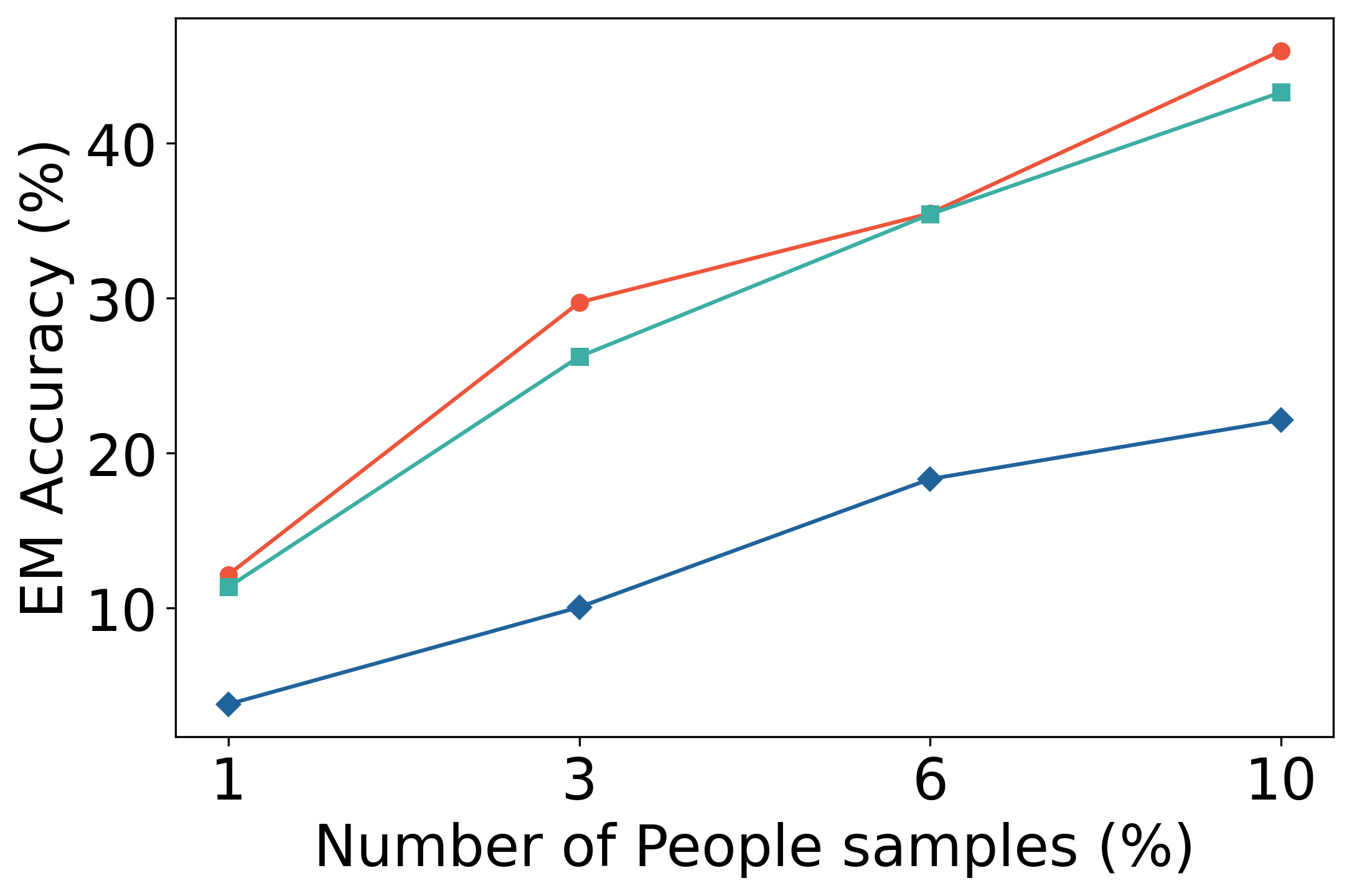}  
\end{minipage}
\begin{minipage}{.33\textwidth}
  \centering
  \includegraphics[width=\linewidth]{img/X2/Recipe.png}  
\end{minipage}
\begin{minipage}{.33\textwidth}
  \centering
  \includegraphics[width=\linewidth]{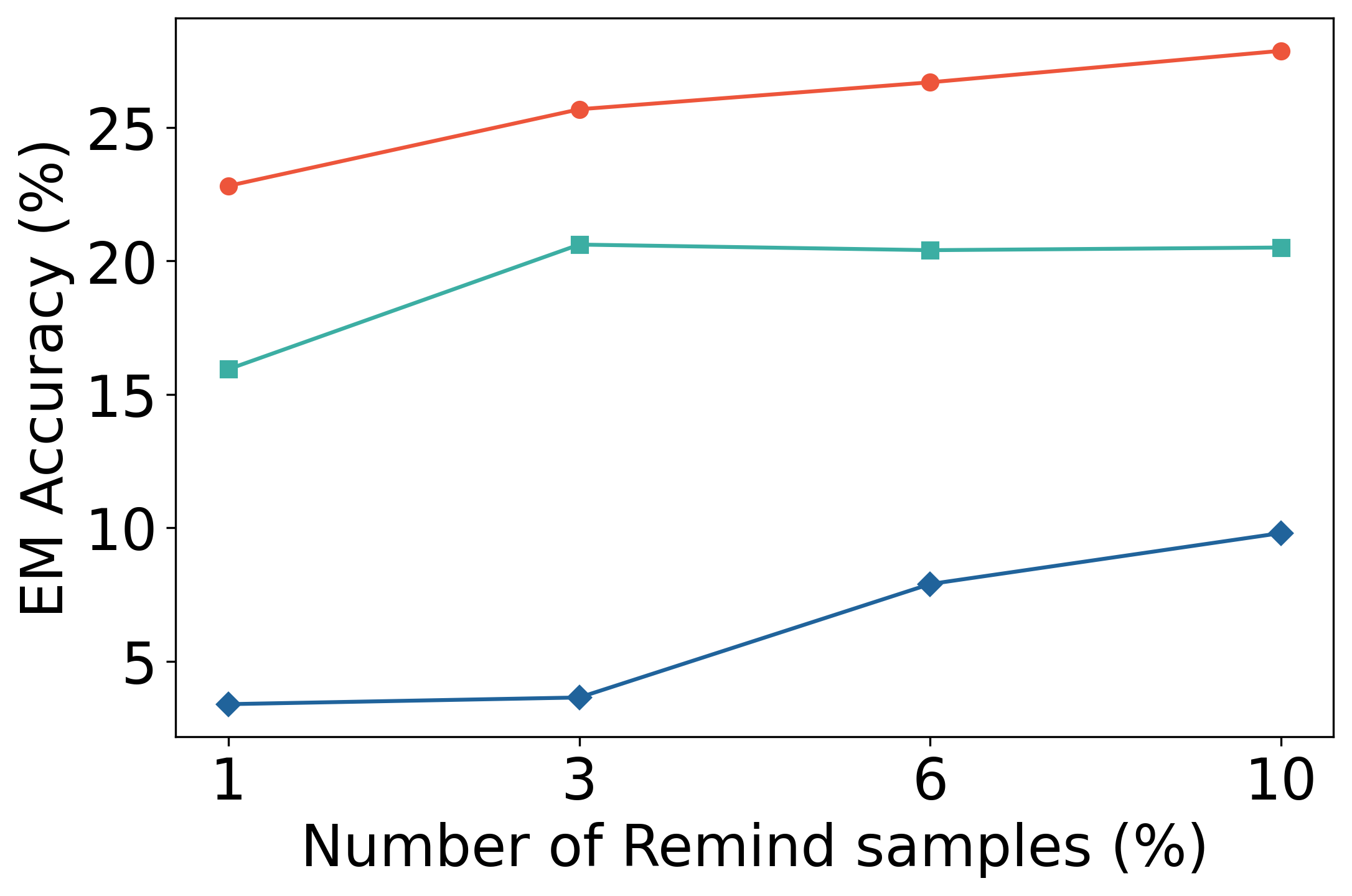}
\end{minipage}
\begin{minipage}{.33\textwidth}
  \centering
  \includegraphics[width=1\linewidth]{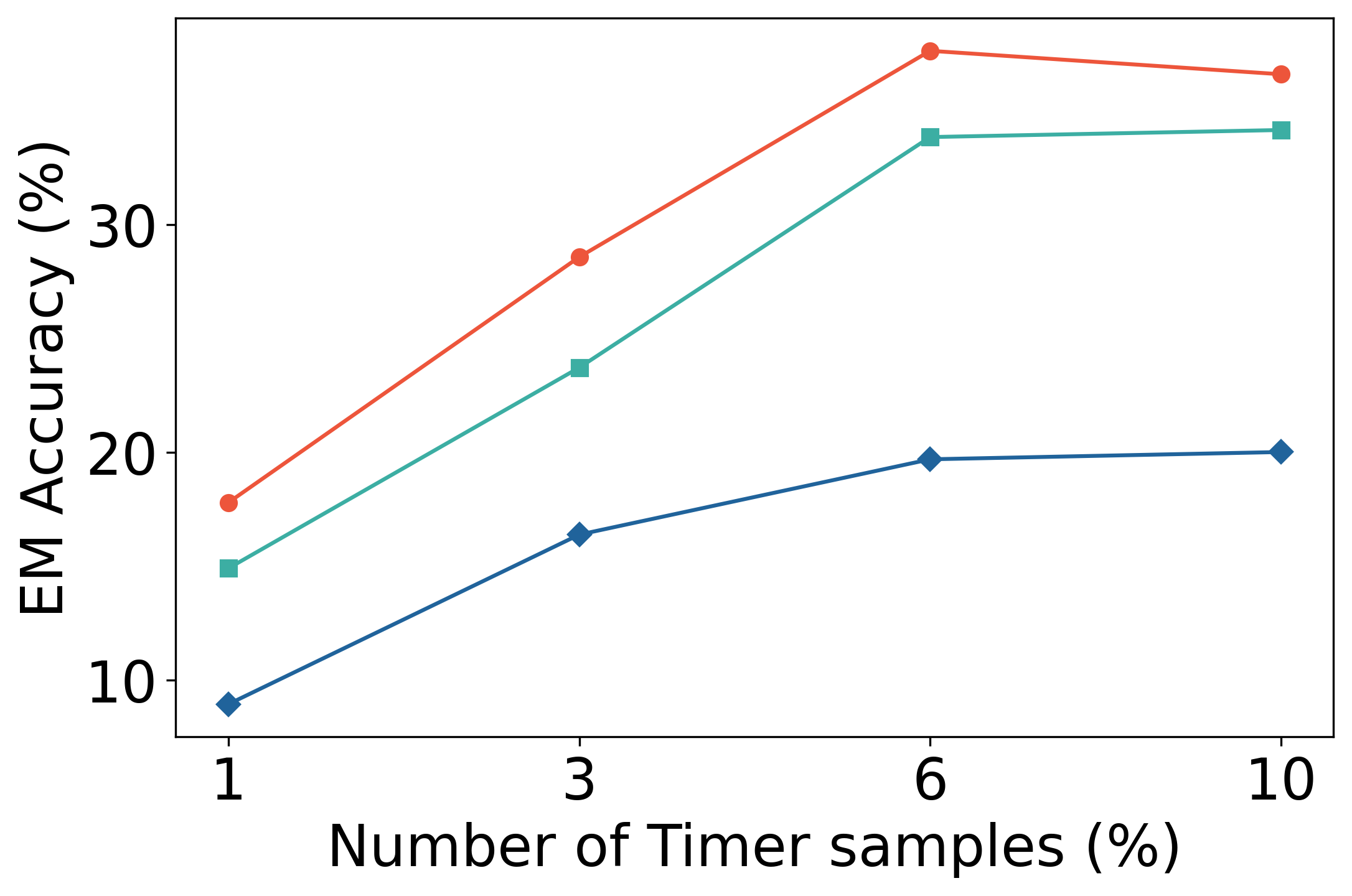}  
\end{minipage}
\begin{minipage}{.33\textwidth}
  \centering
  \includegraphics[width=\linewidth]{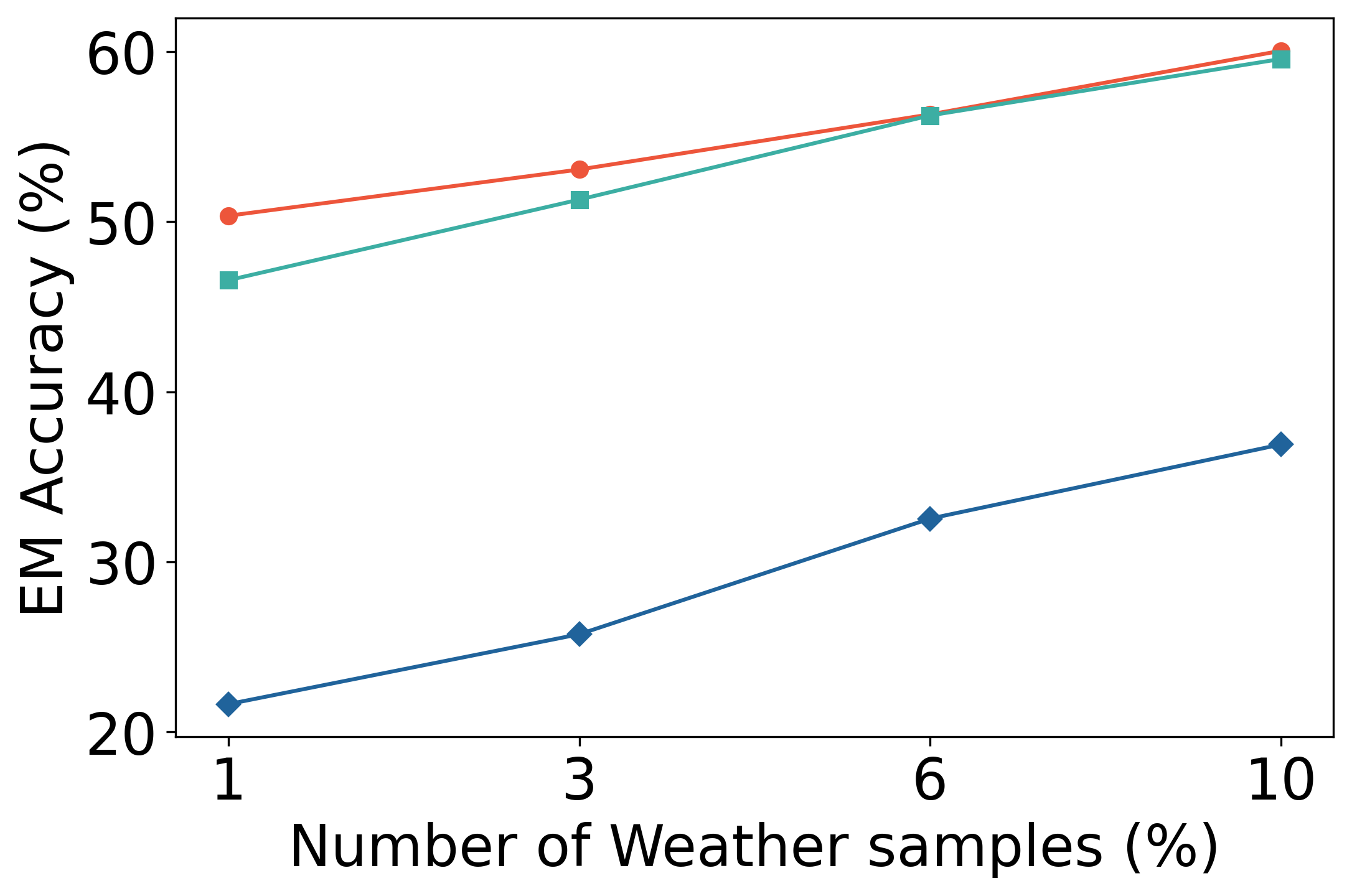}  
\end{minipage}
\caption{Few-shot Exact match accuracies for the \textbf{cross-lingual cross-domain setting} across all 11 target domains. The results are averaged over all target languages.}
\label{fig:results-all-x2}
\end{figure*}

Full few-shot cross-lingual cross-domain results across all 11 target domains are shown in Figure~\ref{fig:results-all-x2} and Tables~\ref{tab:results-complete-x2-spanish}, ~\ref{tab:results-complete-x2-french}, ~\ref{tab:results-complete-x2-german}, ~\ref{tab:results-complete-x2-hindi}, and ~\ref{tab:results-complete-x2-thai}.

\begin{table*}[!ht]
\centering
\renewcommand{\arraystretch}{1.3}
\resizebox{\textwidth}{!}{
\begin{tabular}{ll|ccccccccccc|c}
\Xhline{2\arrayrulewidth}
\textbf{\# Sample} & \textbf{Model} & \textbf{Alarm} & \textbf{Call.} & \textbf{Event} & \textbf{Msg.} & \textbf{Music} & \textbf{News} & \textbf{People} & \textbf{Recipe} & \textbf{Remind} & \textbf{Timer} & \textbf{Weather} & \textbf{Avg} \\ \hline
\multirow{3}{*}{1\%} & Seq2Seq & 45.22 & 33.07 & 34.52 & 22.58 & 10.38 & 20.14 & 12.43 & 11.39 & 18.33 & 20.34 & 48.36 & 25.16 \\
 & NLM & 51.75 & 41.00 & 41.46 & 48.97 & 22.49 & 23.84 & 18.65 & 16.52 & 36.84 & 23.67 & 63.11 & 35.30 \\
 & X2Parser & \textbf{54.94} & \textbf{45.20} & \textbf{41.96} & \textbf{51.91} & \textbf{26.83} & \textbf{31.19} & \textbf{18.83} & \textbf{20.00} & \textbf{42.31} & \textbf{30.80} & \textbf{66.53} & \textbf{39.14} \\ \hline
\multirow{3}{*}{3\%} & Seq2Seq & 52.55 & 50.33 & 39.59 & 30.87 & 16.82 & 30.45 & 31.64 & 14.20 & 23.90 & 30.69 & 58.06 & 34.46 \\
 & NLM & 56.86 & 61.68 & 49.24 & 59.86 & 33.06 & 43.95 & 49.43 & 19.71 & 48.23 & 38.62 & 69.20 & 48.17 \\
 & X2Parser & \textbf{62.36} & \textbf{63.37} & \textbf{52.97} & \textbf{60.70} & \textbf{33.42} & \textbf{54.38} & \textbf{50.47} & \textbf{27.34} & \textbf{52.21} & \textbf{50.58} & \textbf{70.57} & \textbf{52.58} \\ \hline
\multirow{3}{*}{6\%} & Seq2Seq & 63.88 & 58.32 & 46.70 & 45.48 & 25.87 & 36.03 & 42.94 & 21.45 & 37.81 & 34.14 & 63.86 & 43.32 \\
 & NLM & 68.53 & 66.42 & 63.96 & \textbf{70.28} & 45.98 & 56.33 & 59.23 & 28.12 & 52.21 & 58.51 & 75.16 & 58.61 \\
 & X2Parser & \textbf{71.61} & \textbf{68.97} & \textbf{69.54} & 70.09 & \textbf{46.70} & \textbf{59.87} & \textbf{59.70} & \textbf{35.65} & \textbf{56.57} & \textbf{61.70} & \textbf{77.00} & \textbf{61.58} \\ \hline
\multirow{3}{*}{10\%} & Seq2Seq & 67.94 & 64.25 & 61.93 & 50.11 & 32.20 & 43.20 & 52.54 & 34.21 & 46.32 & 44.83 & 73.58 & 51.92 \\
 & NLM & 76.32 & 70.02 & 73.60 & 70.58 & \textbf{56.52} & 58.01 & 67.33 & 50.01 & 57.28 & 64.37 & 80.15 & 65.83 \\
 & X2Parser & \textbf{76.72} & \textbf{73.16} & \textbf{77.33} & \textbf{71.45} & 55.19 & \textbf{64.43} & \textbf{69.77} & \textbf{51.78} & \textbf{58.86} & \textbf{65.98} & \textbf{81.17} & \textbf{67.80} \\ \Xhline{2\arrayrulewidth}
\end{tabular}
}
\caption{Complete results of the \textbf{cross-domain setting}.}
\label{tab:results-complete-cross-domain}
\end{table*}

\begin{table*}[!ht]
\centering
\renewcommand{\arraystretch}{1.3}
\resizebox{\textwidth}{!}{
\begin{tabular}{ll|ccccccccccc|c}
\Xhline{2\arrayrulewidth}
\textbf{\# Sample} & \textbf{Model} & \textbf{Alarm} & \textbf{Call.} & \textbf{Event} & \textbf{Msg.} & \textbf{Music} & \textbf{News} & \textbf{People} & \textbf{Recipe} & \textbf{Remind} & \textbf{Timer} & \textbf{Weather} & \textbf{Avg} \\ \hline
\multirow{3}{*}{1\%} & Seq2Seq & 33.81 & 28.00 & 24.29 & 9.89 & 8.22 & 13.59 & 4.49 & 15.87 & 7.87 & 14.29 & 38.86 & 18.11 \\
 & NLM & 44.41 & 31.75 & 36.53 & 41.95 & 14.36 & 16.34 & 12.82 & \textbf{23.28} & 21.76 & 24.67 & 57.92 & 29.62\\
 & X2Parser & \textbf{51.51} & \textbf{36.67} & \textbf{36.72} & \textbf{51.84} & \textbf{20.86} & \textbf{19.90} & \textbf{15.60} & 19.05 & \textbf{26.16} & \textbf{30.74} & \textbf{59.65} & \textbf{33.52} \\ \hline
\multirow{3}{*}{3\%} & Seq2Seq & 48.58 & 41.75 & 30.51 & 14.07 & 12.35 & 10.68 & 17.31 & 19.84 & 6.94 & 28.57 & 42.82 & 24.86 \\
 & NLM & 53.41 & 50.33 & 43.31 & \textbf{54.37} & \textbf{26.10} & 29.45 & 34.19 & \textbf{24.61} & 25.46 & 38.96 & \textbf{64.03} & 40.38 \\
 & X2Parser & \textbf{56.06} & \textbf{54.75} & \textbf{46.14} & 53.23 & 24.25 & \textbf{33.82} & \textbf{34.61} & 23.54 & \textbf{30.79} & \textbf{44.73} & 63.61 & \textbf{42.32}\\ \hline
 \multirow{3}{*}{6\%} & Seq2Seq & 51.70 & 43.50 & 36.16 & 25.48 & 16.91 & 18.45 & 27.56 & 23.02 & 12.50 & 33.33 & 51.49 & 30.92 \\
 & NLM & 60.32 & 52.83 & 48.02 & 60.84 & \textbf{41.46} & \textbf{47.25} & \textbf{46.58} & 26.19 & 27.70 & 53.10 & 67.41 & 48.34\\
 & X2Parser & \textbf{66.10} & \textbf{61.67} & \textbf{53.30} & \textbf{61.85} & 40.24 & 44.82 & 44.01 & \textbf{27.78} & \textbf{31.48} & \textbf{53.97} & \textbf{68.81} & \textbf{50.37} \\ \hline
 \multirow{3}{*}{10\%} & Seq2Seq &  59.94 & 47.00 & 41.81 & 25.86 & 22.85 & 25.39 & 34.21 & 21.25 & 17.59 & 32.90 & 60.69 & 35.41 \\
 & NLM & 64.57 & 57.08 & 53.11 & 60.08 & 49.86 & \textbf{48.84} & 58.12 & 34.13 & 24.61 & 51.95 & 70.30 & 52.06 \\
 & X2Parser & \textbf{65.81} & \textbf{59.75} & \textbf{54.24} & \textbf{61.98} & \textbf{51.36} & 46.12 & \textbf{59.62} & \textbf{36.51} & \textbf{32.10} & \textbf{56.57} & \textbf{71.45} & \textbf{54.14} \\ \Xhline{2\arrayrulewidth}
\end{tabular}
}
\caption{Complete results of the \textbf{cross-lingual cross-domain setting} in Spanish.}
\label{tab:results-complete-x2-spanish}
\end{table*}

\begin{table*}[!ht]
\centering
\renewcommand{\arraystretch}{1.3}
\resizebox{\textwidth}{!}{
\begin{tabular}{ll|ccccccccccc|c}
\Xhline{2\arrayrulewidth}
\textbf{\# Sample} & \textbf{Model} & \textbf{Alarm} & \textbf{Call.} & \textbf{Event} & \textbf{Msg.} & \textbf{Music} & \textbf{News} & \textbf{People} & \textbf{Recipe} & \textbf{Remind} & \textbf{Timer} & \textbf{Weather} & \textbf{Avg} \\ \hline
\multirow{3}{*}{1\%} & Seq2Seq & 43.94 & 30.77 & 15.72 & 11.55 & 8.07 & 5.66 & 8.09 & 12.15 & 5.87 & 13.16 & 35.67 & 17.33 \\
 & NLM & 53.13 & 35.04 & \textbf{41.51} & 46.75 & 16.07 & 7.55 & 15.32 & 18.78 & 26.01 & 23.51 & 59.74 & 31.22 \\
 & X2Parser & \textbf{54.65} & \textbf{37.48} & \textbf{41.51} & \textbf{49.40} & \textbf{21.67} & \textbf{10.27} & \textbf{18.01} & \textbf{20.44} & \textbf{29.28} & \textbf{26.67} & \textbf{65.07} & \textbf{34.04} \\ \hline
\multirow{3}{*}{3\%} & Seq2Seq & 51.21 & 42.91 & 16.98 & 10.76 & 9.07 & 8.81 & 8.09 & 14.92 & 5.22 & 22.11 & 43.54 & 21.24 \\
 & NLM & 55.66 & 49.58 & 51.57 & \textbf{54.98} & \textbf{25.32} & 17.82 & 33.34 & 23.94 & 30.73 & 34.91 & 65.73 & 40.33\\
 & X2Parser & \textbf{59.70} & \textbf{52.87} & \textbf{54.72} & 52.99 & 24.40 & \textbf{18.03} & \textbf{38.60} & \textbf{27.81} & \textbf{31.74} & \textbf{40.18} & \textbf{65.92} & \textbf{42.45} \\ \hline
 \multirow{3}{*}{6\%} & Seq2Seq & 49.70 & 45.24 & 25.79 & 19.92 & 17.86 & 5.66 & 19.85 & 19.89 & 14.78 & 31.05 & 55.62 & 27.76 \\
 & NLM & 64.08 & 50.00 & 57.86 & \textbf{63.88} & 36.67 & 23.27 & \textbf{48.41} & 28.73 & 28.48 & 50.18 & \textbf{74.72} & 47.84\\
 & X2Parser & \textbf{66.77} & \textbf{59.22} & \textbf{58.49} & 59.49 & \textbf{38.45} & \textbf{25.16} & 46.81 & \textbf{29.84} & \textbf{35.29} & \textbf{55.62} & 71.82 & \textbf{49.72} \\ \hline
 \multirow{3}{*}{10\%} & Seq2Seq & 50.00 & 46.34 & 27.67 & 29.48 & 24.29 & 10.18 & 25.43 & 25.10 & 16.09 & 28.42 & 57.81 & 30.98 \\
 & NLM & 59.64 & 52.68 & \textbf{61.22} & \textbf{62.29} & 48.93 & 22.59 & 57.35 & 31.49 & 32.75 & \textbf{52.81} & \textbf{76.69} & 50.77\\
 & X2Parser & \textbf{62.32} & \textbf{56.84} & 58.70 & 61.89 & \textbf{49.17} & \textbf{25.58} & \textbf{60.54} & \textbf{38.49} & \textbf{36.31} & 52.28 & 75.37 & \textbf{52.50} \\ \Xhline{2\arrayrulewidth}
\end{tabular}
}
\caption{Complete results of the \textbf{cross-lingual cross-domain setting} in French.}
\label{tab:results-complete-x2-french}
\end{table*}

\begin{table*}[!ht]
\centering
\renewcommand{\arraystretch}{1.3}
\resizebox{\textwidth}{!}{
\begin{tabular}{ll|ccccccccccc|c}
\Xhline{2\arrayrulewidth}
\textbf{\# Sample} & \textbf{Model} & \textbf{Alarm} & \textbf{Call.} & \textbf{Event} & \textbf{Msg.} & \textbf{Music} & \textbf{News} & \textbf{People} & \textbf{Recipe} & \textbf{Remind} & \textbf{Timer} & \textbf{Weather} & \textbf{Avg} \\ \hline
\multirow{3}{*}{1\%} & Seq2Seq & 35.13 & 21.61 & 7.36 & 8.81 & 7.98 & 11.74 & 3.58 & 6.90 & 0.89 & 17.19 & 24.15 & 13.21 \\
 & NLM & 46.27 & 37.43 & 44.58 & 27.81 & 18.95 & 24.50 & 14.70 & 11.84 & 16.03 & 22.00 & 57.86 & 29.27\\
 & X2Parser & \textbf{52.03} & \textbf{39.32} & \textbf{45.40} & \textbf{34.72} & \textbf{21.68} & \textbf{33.97} & \textbf{15.17} & \textbf{15.29} & \textbf{19.98} & \textbf{27.09} & \textbf{60.52} & \textbf{33.20} \\ \hline
\multirow{3}{*}{3\%} & Seq2Seq & 38.81 & 35.45 & 15.34 & 8.81 & 13.89 & 13.26 & 15.41 & 11.03 & 2.46 & 28.12 & 23.69 & 18.75 \\
 & NLM & 52.50 & 48.44 & 57.87 & 38.86 & \textbf{28.94} & 38.38 & 38.35 & 16.32 & 21.55 & 36.98 & 64.84 & 40.28\\
 & X2Parser & \textbf{52.69} & \textbf{50.18} & \textbf{58.28} & \textbf{39.90} & 25.82 & \textbf{48.36} & \textbf{45.04} & \textbf{24.71} & \textbf{23.64} & \textbf{39.32} & \textbf{65.83} & \textbf{43.07} \\ \hline
 \multirow{3}{*}{6\%} & Seq2Seq & 37.68 & 38.10 & 20.25 & 17.62 & 15.80 & 18.18 & 25.81 & 13.10 & 6.49 & 30.47 & 36.67 & 23.65 \\
 & NLM & 54.77 & 50.00 & 62.78 & 42.14 & 34.75 & 47.35 & 50.54 & 21.72 & 21.92 & 47.27 & 70.31 & 45.78\\
 & X2Parser & \textbf{59.39} & \textbf{55.31} & \textbf{69.32} & \textbf{48.70} & \textbf{35.08} & \textbf{51.77} & \textbf{53.53} & \textbf{28.74} & \textbf{25.28} & \textbf{50.00} & \textbf{73.20} & \textbf{50.03} \\ \hline
 \multirow{3}{*}{10\%} & Seq2Seq & 39.38 & 35.90 & 22.09 & 13.47 & 25.17 & 22.83 & 30.24 & 17.83 & 7.38 & 32.03 & 36.89 & 25.75 \\
 & NLM & \textbf{60.13} & 54.76 & 64.01 & 46.98 & \textbf{42.16} & 46.84 & 57.35 & 40.12 & 22.30 & 49.22 & 73.88 & 50.70\\
 & X2Parser & 56.75 & \textbf{56.23} & \textbf{68.92} & \textbf{49.05} & 40.96 & \textbf{52.65} & \textbf{65.59} & \textbf{42.64} & \textbf{24.91} & \textbf{51.56} & \textbf{75.47} & \textbf{53.16} \\ \Xhline{2\arrayrulewidth}
\end{tabular}
}
\caption{Complete results of the \textbf{cross-lingual cross-domain setting} in German.}
\label{tab:results-complete-x2-german}
\end{table*}

\begin{table*}[!ht]
\centering
\renewcommand{\arraystretch}{1.3}
\resizebox{\textwidth}{!}{
\begin{tabular}{ll|ccccccccccc|c}
\Xhline{2\arrayrulewidth}
\textbf{\# Sample} & \textbf{Model} & \textbf{Alarm} & \textbf{Call.} & \textbf{Event} & \textbf{Msg.} & \textbf{Music} & \textbf{News} & \textbf{People} & \textbf{Recipe} & \textbf{Remind} & \textbf{Timer} & \textbf{Weather} & \textbf{Avg} \\ \hline
\multirow{3}{*}{1\%} & Seq2Seq & 11.99 & 13.30 & 9.85 & 1.46 & 1.65 & 3.23 & 1.91 & 3.08 & 0.00 & 0.00 & 6.89 & 4.85 \\
 & NLM & \textbf{16.10} & 17.67 & 15.15 & 20.39 & 9.92 & 12.73 & \textbf{9.54} & 5.73 & 1.27 & \textbf{3.25} & 30.44 & 12.92 \\
 & X2Parser & \textbf{16.10} & \textbf{25.04} & \textbf{25.00} & \textbf{20.55} & \textbf{10.60} & \textbf{15.59} & 7.76 & \textbf{7.64} & \textbf{12.88} & \textbf{3.25} & \textbf{31.63} & \textbf{16.00}\\ \hline
\multirow{3}{*}{3\%} & Seq2Seq & 11.61 & 18.00 & 14.39 & 6.80 & 4.83 & 3.76 & 5.73 & 3.96 & 0.38 & 2.56 & 14.07 & 7.83 \\
 & NLM & 14.23 & 28.62 & 23.49 & \textbf{24.27} & 13.54 & 14.87 & 16.29 & 9.25 & 3.28 & 5.47 & \textbf{34.33} & 17.06\\
 & X2Parser & \textbf{25.59} & \textbf{34.76} & \textbf{40.91} & 21.85 & \textbf{14.46} & \textbf{34.77} & \textbf{19.08} & \textbf{17.03} & \textbf{15.66} & \textbf{16.93} & 32.64 & \textbf{24.88} \\ \hline
 \multirow{3}{*}{6\%} & Seq2Seq & 14.61 & 13.73 & 18.18 & 6.31 & 10.26 & 4.84 & 11.83 & 5.29 & 0.38 & 2.56 & 12.87 & 9.17 \\
 & NLM & \textbf{29.63} & \textbf{35.29} & 29.80 & 22.17 & \textbf{20.11} & 21.33 & 20.74 & 9.84 & 2.15 & 16.07 & \textbf{38.42} & 22.32\\
 & X2Parser & 27.97 & 34.05 & \textbf{47.47} & \textbf{23.62} & 17.36 & \textbf{36.74} & \textbf{21.88} & \textbf{16.59} & \textbf{16.54} & \textbf{26.84} & 33.34 & \textbf{27.49} \\ \hline
 \multirow{3}{*}{10\%} & Seq2Seq & 11.24 & 22.53 & 18.94 & 8.74 & 13.31 & 7.23 & 13.54 & 6.52 & 0.38 & 5.64 & 19.07 & 11.56 \\
 & NLM & 22.58 & 25.61 & 32.07 & 21.36 & \textbf{24.38} & 20.25 & 24.30 & 15.13 & 2.40 & 14.19 & \textbf{38.61} & 21.90\\
 & X2Parser & \textbf{30.71} & \textbf{43.92} & \textbf{50.25} & \textbf{22.33} & 22.73 & \textbf{33.33} & \textbf{25.06} & \textbf{22.02} & \textbf{16.54} & \textbf{20.17} & 35.53 & \textbf{29.33} \\ \Xhline{2\arrayrulewidth}
\end{tabular}
}
\caption{Complete results of the \textbf{cross-lingual cross-domain setting} in Hindi.}
\label{tab:results-complete-x2-hindi}
\end{table*}

\begin{table*}[!ht]
\centering
\renewcommand{\arraystretch}{1.3}
\resizebox{\textwidth}{!}{
\begin{tabular}{ll|ccccccccccc|c}
\Xhline{2\arrayrulewidth}
\textbf{\# Sample} & \textbf{Model} & \textbf{Alarm} & \textbf{Call.} & \textbf{Event} & \textbf{Msg.} & \textbf{Music} & \textbf{News} & \textbf{People} & \textbf{Recipe} & \textbf{Remind} & \textbf{Timer} & \textbf{Weather} & \textbf{Avg} \\ \hline
\multirow{3}{*}{1\%} & Seq2Seq & 7.82 & 10.81 & 3.40 & 0.00 & 1.82 & 0.64 & 0.94 & 6.90 & 2.31 & 0.00 & 2.59 & 3.38 \\
 & NLM & \textbf{37.08} & 26.27 & 10.20 & 16.92 & 9.93 & 4.46 & \textbf{4.40} & 8.74 & 14.65 & 1.06 & 26.82 & 14.59\\
 & X2Parser & 34.12 & \textbf{26.77} & \textbf{16.33} & \textbf{18.95} & \textbf{11.29} & \textbf{8.92} & 4.25 & \textbf{13.33} & \textbf{25.76} & \textbf{1.24} & \textbf{34.87} & \textbf{17.80}\\ \hline
\multirow{3}{*}{3\%} & Seq2Seq & 12.93 & 12.96 & 4.08 & 1.02 & 5.08 & 0.64 & 3.77 & 7.59 & 3.20 & 0.53 & 4.60 & 5.13 \\
 & NLM & 30.61 & 31.54 & 16.49 & \textbf{24.03} & \textbf{12.75} & 3.82 & 8.96 & 16.78 & 22.02 & \textbf{2.29} & 27.59 & 17.90\\
 & X2Parser & \textbf{35.03} & \textbf{36.09} & \textbf{30.16} & 20.14 & 10.07 & \textbf{12.32} & \textbf{11.32} & \textbf{19.08} & \textbf{26.55} & 1.77 & \textbf{37.36} & \textbf{21.81}\\ \hline
 \multirow{3}{*}{6\%} & Seq2Seq & 6.46 & 16.95 & 4.76 & 2.03 & 6.86 & 1.27 & 6.60 & 5.52 & 5.32 & 1.06 & 6.03 & 5.71 \\
 & NLM & 34.39 & 35.31 & 15.42 & \textbf{29.10} & 10.48 & 5.73 & 10.85 & 20.46 & 21.73 & \textbf{2.65} & 30.36 & 19.68\\
 & X2Parser & \textbf{35.83} & \textbf{38.21} & \textbf{30.16} & 25.21 & \textbf{16.19} & \textbf{11.68} & \textbf{11.16} & \textbf{20.92} & \textbf{24.87} & 1.77 & \textbf{34.36} & \textbf{22.76}\\ \hline
 \multirow{3}{*}{10\%} & Seq2Seq & 10.88 & 16.53 & 2.04 & 4.57 & 10.45 & 2.91 & 7.25 & 9.90 & 7.56 & 1.06 & 10.02 & 7.56 \\
 & NLM & \textbf{35.75} & 26.34 & 12.70 & \textbf{25.89} & 15.92 & 4.46 & \textbf{19.34} & 21.84 & 20.45 & \textbf{2.65} & 38.39 & 20.34 \\
 & X2Parser & 28.00 & \textbf{39.75} & \textbf{34.01} & 24.70 & \textbf{22.04} & \textbf{16.56} & 19.02 & \textbf{25.29} & \textbf{29.50} & 2.47 & \textbf{42.43} & \textbf{25.80}\\ \Xhline{2\arrayrulewidth}
\end{tabular}
}
\caption{Complete results of the \textbf{cross-lingual cross-domain setting} in Thai.}
\label{tab:results-complete-x2-thai}
\end{table*}

\end{document}